\documentclass[lettersize,journal]{IEEEtran}
\usepackage{amsmath,amsfonts}
\usepackage{array}
\usepackage[caption=false,font=normalsize,labelfont=sf,textfont=sf]{subfig}
\usepackage{textcomp}
\usepackage{stfloats}
\usepackage{url}
\usepackage{verbatim}
\usepackage{graphicx}
\usepackage{cite}
\usepackage{amsthm}
\usepackage{amsmath}
\usepackage{multirow}
\usepackage{mathtools}
\usepackage{booktabs}
\usepackage{fontawesome5}
\usepackage[table]{xcolor}
\usepackage{color}
\usepackage{pifont}
\usepackage{bbding}
\usepackage{algorithm}
\usepackage{algpseudocode}
\usepackage{amssymb} 
\begin{document}

\title{Informative Sample Selection Model for Skeleton-based Action Recognition with Limited Training Samples}

\author{Zhigang Tu,~\IEEEmembership{Senior Member,~IEEE}, ~Zhengbo Zhang~\IEEEmembership{Member,~IEEE}, ~Jia Gong, ~Junsong Yuan,~\IEEEmembership{Fellow,~IEEE}, ~~Bo Du,~\IEEEmembership{Senior Member,~IEEE}%
\thanks{Corresponding author: Zhengbo  Zhang (e-mail: zhangzb@whu.edu.cn)}
\thanks{Zhigang Tu is with the State Key Laboratory of Information Engineering in Surveying, Mapping and Remote Sensing, Wuhan University, Wuhan 430072, China (e-mail: tuzhigang@whu.edu.cn).}%
\thanks{Zhengbo Zhang is with the Information Systems Technology and Design Pillar, Singapore University of Technology and Design, Singapore 487372.}%
\thanks{Jia Gong is with the Shanghai Academy of AI for Science, Shanghai, China.}%
\thanks{Junsong Yuan is with the Department of Computer Science and Engineering, University at Buffalo, State University of New York, NY 14228 USA.}%
\thanks{Bo Du is with the School of Computer Science, Wuhan University, Wuhan 430072, China.}
}

\markboth{IEEE Transactions on Image Processing}%
{Shell \MakeLowercase{\textit{et al.}}: A Sample Article Using IEEEtran.cls for IEEE Journals}

\newcommand{\zb}[1]{\textcolor[rgb]{0.78, 0.2274, 0.333}{\textbf{#1}}}
\definecolor{yellow}{rgb}{1, 1, 0.7}
\definecolor{Second}{rgb}{1, 0.85, 0.7}
\definecolor{Best}{rgb}{1, 0.7, 0.7}
\newcommand{\refine}[1]{\textcolor{red}{#1}}

\maketitle

\begin{abstract}
Skeleton-based human action recognition aims to classify human skeletal sequences, which are spatiotemporal representations of actions, into predefined categories. To reduce the reliance on costly annotations of skeletal sequences while maintaining competitive recognition accuracy, the task of 3D Action Recognition with Limited Training Samples, also known as semi-supervised 3D Action Recognition, has been proposed. In addition, active learning, which aims to proactively select the most informative unlabeled samples for annotation, has been explored in semi-supervised 3D Action Recognition for training sample selection.
Specifically, researchers adopt an encoder–decoder framework to embed skeleton sequences into a latent space, where clustering information, combined with a margin-based selection strategy using a multi-head mechanism, is utilized to identify the most informative sequences in the unlabeled set for annotation. However, the most representative skeleton sequences may not necessarily be the most informative for the action recognizer, as the model may have already acquired similar knowledge from previously seen skeleton samples. As a result, selecting such samples with limited new information to train the action recognizer may lead to sub-optimal recognition performance.
To address this limitation, we reformulate Semi-supervised 3D action recognition via active learning from a novel perspective by casting it as a Markov Decision Process (MDP).
Built upon the well-established MDP framework and its training paradigm, we train an informative sample selection model to intelligently guide the selection of skeleton sequences for annotation.
To enhance the representational capacity of the factors in the state-action pairs within our MDP framework, we project them from Euclidean space to hyperbolic space. Furthermore, we introduce a meta tuning strategy based on meta-learning to accelerate the deployment of our method in real-world scenarios. Extensive experiments on three 3D action recognition benchmarks demonstrate the effectiveness of our method.
\end{abstract}


\section{Introduction}

\IEEEPARstart{H}{uman}  action recognition~\cite{wang2018action,ye2020dynamic,journals/tip/ZhuSLZL23,journals/tip/ShiZCL20,journals/tip/XuYZWLJ19} aims to classify actions from spatiotemporal sequences into a set of pre-defined categories, and serves as a core component in various applications such as activity understanding~\cite{chang2020clustering}, interaction modeling~\cite{liu2016spatio}, and robotic control~\cite{rodomagoulakis2016multimodal}. Based on the type of input data, existing action recognition methods can be broadly categorized into RGB-based~\cite{yue2022action,journals/tip/LinDHZ23,journals/tip/XuYZWLJ19,journals/tip/0001LZDLY19}, depth-based~\cite{chen2014improving}, and 3D skeleton-based approaches~\cite{xu2019semisupervised,journals/tip/GuanYHFL24,journals/tip/MyungSXW24}. Among these data, 3D skeleton data, which represents the human body as a set of keypoints in 3D space, has received increasing attention in recent years~\cite{shahroudy2016ntu,liu2016spatio}. This is because, compared to RGB or depth data, skeleton representations offer compact, high-level descriptions of human motion and are generally more robust to appearance variations, background clutter, and viewpoint changes~\cite{zhang2019comprehensive}. Furthermore, skeleton sequences can be efficiently captured by commodity depth sensors, enabling the development of numerous supervised methods to learn spatiotemporal features for skeleton-based action recognition~\cite{xu2022skeleton}.

\begin{figure}[!t]
\centering
\includegraphics[width=3.5in]{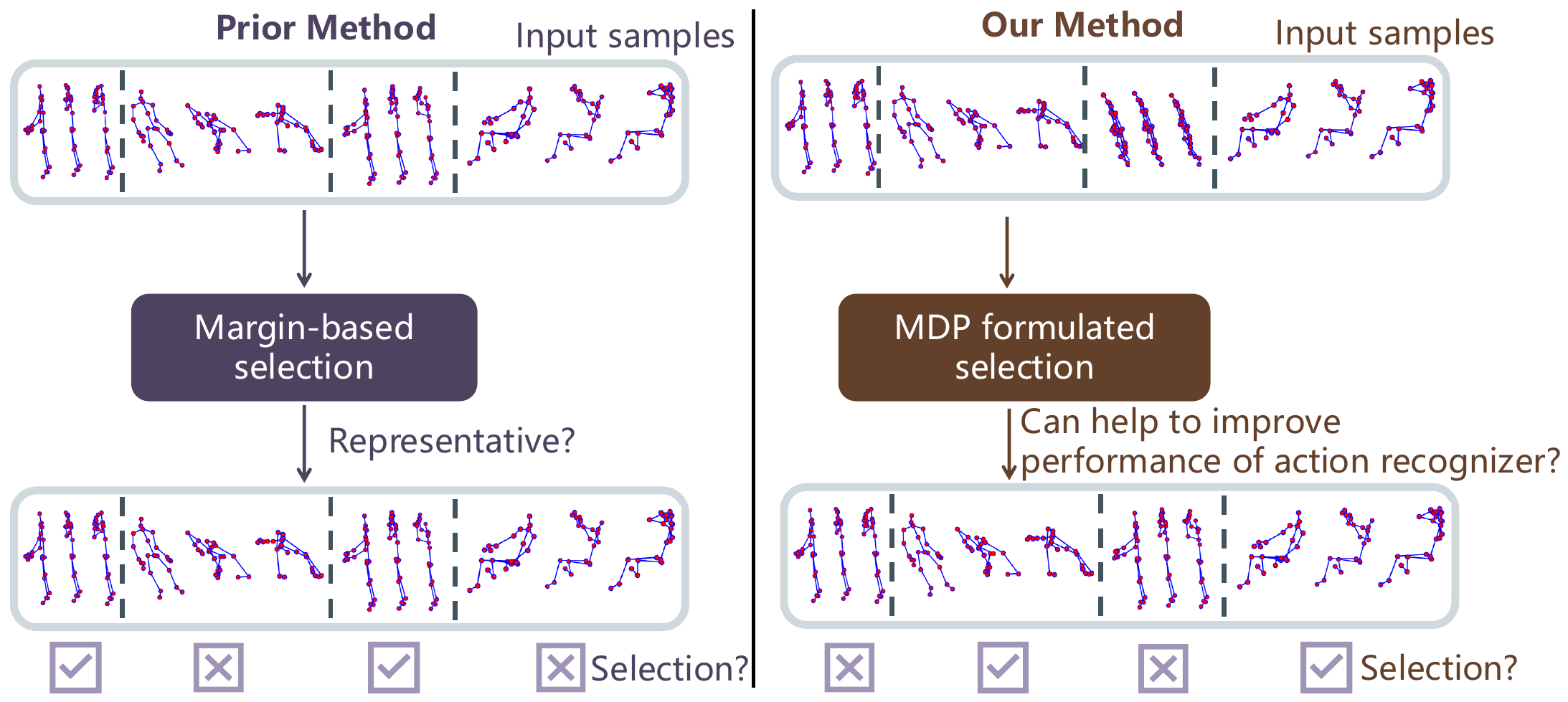}
\caption{Motivation of our method.
Previous semi-supervised 3D action recognition via active learning (S3ARAL) approaches~\cite{li2023sar} rely on margin-based selection strategy that aims to identify representative samples. However, such strategy may select samples with limited novel information, leading to sub-optimal performance of the trained action recognizer. In contrast, we propose a novel perspective by reformulating S3ARAL as a Markov Decision Process (MDP)~\cite{mnih2015human}. Within this framework, we train an informative sample selection model to intelligently choose training samples that are more likely to improve the action recognizer’s performance, thereby enabling the training of a more effective model.}
\label{fig:motivation}
\end{figure}

Recently, with the rapid development of deep learning~\cite{ke2017new,du2015hierarchical,journals/tip/0001LZDLY19, zhang2022distilling}, deep neural networks~\cite{journals/tip/ZhangSHS18,journals/tip/ZhangCLWDo16,zhang2025,tu2019action,tu2023dtcm,liu2022motion} have been extensively explored for modeling spatio-temporal representations of skeleton sequences in supervised settings. Specifically, Recurrent Neural Networks  have been applied to capture temporal dependencies in action sequences~\cite{du2015hierarchical}, while Convolutional Neural Networks  have been adopted by transforming joint coordinates into 2D maps~\cite{wang2018action}. Besides, Graph Convolutional Networks have gained popularity due to their strong performance in modeling the structured topology of skeleton data~\cite{ye2020dynamic}. Despite their promising results, these methods typically rely on large-scale annotated datasets for training~\cite{du2015hierarchical}. However, obtaining accurate frame-level annotations is often labor-intensive and requires annotators to possess expert knowledge of human skeletons, which poses significant challenges to scalability and generalization across unseen actions or subjects~\cite{li2023sar}.

To reduce the reliance on extensive annotations without sacrificing recognition accuracy in skeleton-based action recognition, semi-supervised learning approaches have been proposed, and received considerable attention.
Specifically, 
the semi-supervised setting assumes that annotated auxiliary classes are unavailable. Methods such as ASSL~\cite{si2020adversarial}, MS$^2$L~\cite{lin2020ms2l}, and SC3D~\cite{thoker2021skeleton} aim to learn informative representations from both labeled and unlabeled data while preserving classification performance.
Nevertheless, these methods typically overlook the fact that not all labeled samples contribute equally to model training. Selecting the representative samples for annotation can enhance classifier effectiveness while further reducing annotation cost~\cite{li2023sar}.

To address this issue, the previous work explores AL-SAR by incorporating active learning (AL), which aims to proactively select the most informative unlabeled samples for annotation, with semi-supervised 3D action recognition~\cite{li2023sar}.  Specifically, AL-SAR first employs an encoder–decoder framework to encode skeleton sequences into a latent space. It then leverages the clustering information in this latent space, together with a margin-based selection strategy built on a multi-head mechanism, to identify informative skeleton sequences  in the unlabeled set for annotation.
However, the representative skeleton sequences may not always be the most informative for the action recognizer, as the model may have already learned similar patterns from previously labeled samples. Consequently, selecting such redundant samples may offer limited benefit and lead to sub-optimal performance for the action recognizer.
Hence, it is crucial to boost Semi-supervised 3D Action Recognition via AL (S3ARAL) in a more intelligent and effective manner.

In this paper, inspired by \cite{Gong2022MetaAT}, we handle this problem from \emph{a novel perspective} by formulating S3ARAL as a Markov Decision Process (MDP)~\cite{mnih2015human} (see Figure \ref{fig:motivation}). Within this MDP framework, the action corresponds to selecting the representative samples for annotation.
Our key insight of such formulation is that MDP and S3ARAL share a common objective of maximizing long-term benefits: MDP aims to maximize expected cumulative rewards, while S3ARAL seeks to maximize action recognizer's performance by selecting the  informative samples for annotation. Moreover, MDP offers solid theoretical foundations~\cite{mnih2015human} and has demonstrated its capability to make informed decisions (action) in complex state spaces across multiple domains, like robotics~\cite{van2016deep} and Game AI~\cite{kulkarni2016hierarchical}. Based on our formulation, we set the reward in the MDP as the action recognizer's performance improvement among training stages, with the objective of maximizing cumulative performance gain of the recognizer. 

In addition, human skeleton data naturally form tree-like graphs, where hierarchical relationships exist among joints. As a result, directly computing features of the skeleton data in the Euclidean space may fail to capture these structural properties effectively. Motivated by the exponential volume growth of hyperbolic space~\cite{qu2024llms}, which enables efficient representation of hierarchical skeleton data, we project the factors in the state-action space of our MDP formulation from the Euclidean space to the hyperbolic space to enhance their representational capacity.
Furthermore, when our method is deployed in real-world scenarios, it typically requires re-training on the enlarged labeled dataset, which can be time-consuming. To address this issue, we design a meta-tuning strategy based on meta learning~\cite{finn2017model} to improve the generalizability of the proposed method, allowing a rapid deployment.

To ensure a fair comparison with the previous S3ARAL method~\cite{li2023sar}, we follow its experimental setup and conduct experiments on three commonly used 3D action recognition datasets: UWA3D Multiview Activity II~\cite{Rahmani2014HOPCHO}, North-Western UCLA~\cite{Wang2014CrossViewAM}, and NTU RGB+D 60~\cite{Shahroudy2016NTURA}. 
Extensive experiments on these datasets validate the effectiveness of our method.

The main contributions are summarized as follows:
\begin{itemize}
\item Existing S3ARAL method often overlooks whether the selected samples are informative for the action recognizer, leading to sub-optimal performance of the trained model. To address this issue, we approach the task from a new perspective by formulating it as  a MDP, and design a novel training framework that directly links the improvement of the recognizer’s performance with the informativeness of the selected samples, enabling the training of a more powerful action recognizer.

\item To enhance the representational capacity of factors in the state-action pairs of our MDP framework, we map them from the Euclidean space to the hyperbolic space. Besides, we introduce a meta tuning strategy to accelerate the deployment of our method in real-world scenarios.

\item We conduct extensive experiments on three 3D action recognition datasets, where the results show that our method effectively selects informative samples, enabling the training of a high-performing action recognizer. Moreover, our method exhibits strong generalization ability and consistently achieves state-of-the-art performance across all three datasets.
\end{itemize}

\section{Related Work}

\subsection{Human Action Recognition}
Existing human action recognition methods~\cite{journals/tip/ZhangWWQW18,journals/tip/ZhuSLZL23,journals/tip/ShiZCL20,journals/tip/LinDHZ23,journals/tip/XuYZWLJ19,journals/tip/GuanYHFL24,journals/tip/MyungSXW24} can be broadly categorized into two groups: RGB-based methods~\cite{journals/tip/LinDHZ23,journals/tip/XuYZWLJ19} and skeleton-based methods~\cite{journals/tip/GuanYHFL24,journals/tip/MyungSXW24,journals/tip/ZhuSLZL23,journals/tip/ShiZCL20}. Due to the compact yet informative nature of skeleton sequences for representing human behavior, skeleton-based action recognition has attracted substantial research interest.

Early skeleton-based action recognition methods explore a variety of network architectures to process skeleton sequences. For instance, Liu et al.~\cite{liu2016spatio} introduce a spatial-temporal LSTM for skeleton-based action recognition, while Ke et al.~\cite{ke2017new} propose converting skeleton sequences into grayscale images, which are then fed into a CNN, benefiting from the advances in deep learning~\cite{zhang2024Diff,yan2018spatial,li2019actional,wang20233mformer,zhang2025visual}. Over time, graph convolutional networks (GCNs) have emerged as a dominant architecture for this task. Yan et al.~\cite{yan2018spatial} pioneer this direction with ST-GCN, the first GCN-based framework for skeleton-based action recognition. Building upon this, various enhanced GCN models have been proposed, including AS-GCN~\cite{li2019actional},  and CTR-GCN~\cite{chen2021channel}. More recently, transformer-based models have gained popularity for this task. Several works~\cite{wang20233mformer,foo2023unified} have proposed end-to-end transformer architectures for skeleton-based action recognition, such as the 3Mformer \cite{wang20233mformer} and the UPS \cite{foo2023unified}.

Different from these methods, this work tackles a relatively new task: Semi-supervised 3D Action Recognition via Active Learning. It selects informative human sequence data from the unlabeled set through active learning and trains an action recognizer based on the annotated samples.

\subsection{Semi-supervised Human Action Recognition}
Semi-supervised learning for action recognition aims to extract motion representations from unlabeled data by regularizing features or solving pretext tasks. Many studies have proposed semi-supervised methods tailored for unlabeled RGB data~\cite{Zhang2017SemiSupervisedIA,Singh2021SemiSupervisedAR,Chang2020ClusteringDD}. Specifically, \cite{Zhang2017SemiSupervisedIA} focus on predicting motion flows to learn video representations, while~\cite{Chang2020ClusteringDD} introduces a convolutional auto-encoder that separately captures spatial and temporal information from raw videos. Leveraging the observation that varying frame rates do not alter the semantic meaning of actions,~\cite{Singh2021SemiSupervisedAR} designs a two-stream contrastive model in the temporal domain to utilize unlabeled videos at different playback speeds.

Although these methods achieve promising results on RGB video data, they are less suitable for long-term skeleton sequences. Unlike RGB videos, which contain rich appearance and scene context, skeleton data primarily encodes human motion. Thus, effectively capturing joint and bone movement patterns is critical for semi-supervised skeleton action recognition. To this end, Zheng et al.~\cite{Zheng2018UnsupervisedRL} proposes a joint inpainting approach to learn features from unlabeled skeleton sequences, followed by the work of Si et al.~\cite{Si2020AdversarialSL}. However, such self-supervised methods, which rely on inpainting key joints, fail to capture holistic motion dynamics and are not well-suited for GCN-based architectures. Lin et al.~\cite{Lin2020MS2LMS} introduce a multi-task self-supervised framework for skeleton data, but their model struggled to extract joint-bone fused features using a network structure optimized for action recognition. Moreover,~\cite{Zheng2018UnsupervisedRL,Si2020AdversarialSL,Lin2020MS2LMS} still adopt RNN-based backbones, which are suboptimal for capturing spatial-temporal dependencies.

In this paper, unlike previous semi-supervised skeleton-based action recognition methods that often randomly select a batch of samples for annotation to train the action recognizer, we adopt active learning to select informative skeleton samples for annotation.

\begin{figure*}[!ht]
\centering
\includegraphics[width=7.2in]{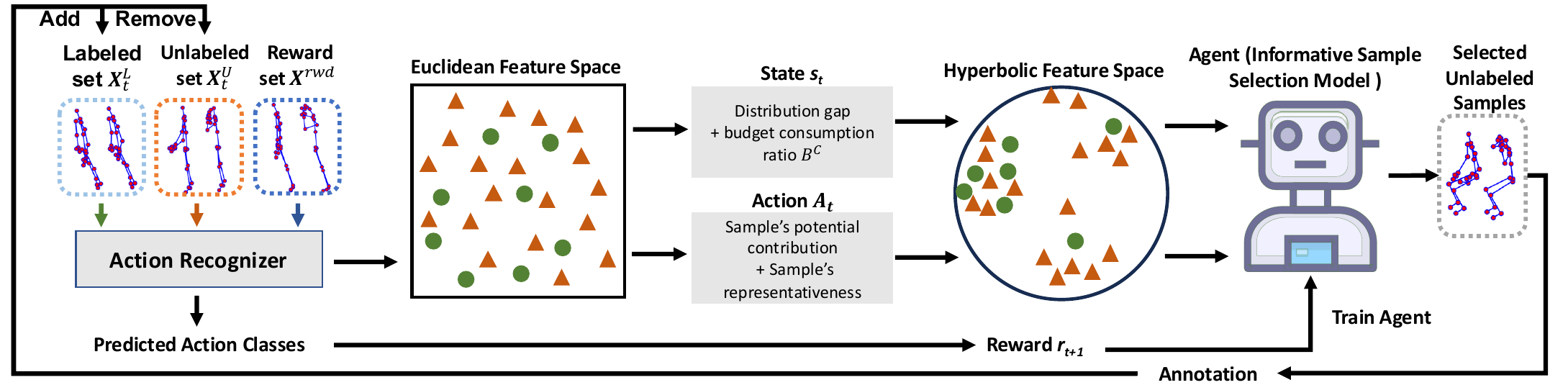}
\caption{The overall pipeline of our method. In the task of semi-supervised 3D action recognition via active learning, given an unlabeled dataset and an annotation budget, we divide the dataset into a labeled set, an unlabeled set, and a reward set. We then design an Informative Sample Selection Model (ISSM) to select informative samples for annotation. The annotated samples are used to train the action recognizer.
To ensure that the ISSM receives sufficient information for effective sample selection, we carefully construct the state-action pairs. The state encodes the distribution gap between the labeled and unlabeled sets, along with the budget consumption ratio. The action captures both the sample’s potential contribution to improving the action recognizer and its representativeness.
To enhance the expressiveness of the state-action representations for the skeleton-based action recognition task, we project them from the Euclidean space to hyperbolic space. This is motivated by the exponential volume growth of hyperbolic space, which is well-suited for modeling the hierarchical structure of human skeletons. The performance improvement of the action recognizer between consecutive iterations is treated as the reward for training the ISSM.
}
\label{fig:pipeline}
\end{figure*}

\subsection{Markov Decision Process}
Markov Decision Process (MDP) has become the standard model for studying how agents make optimal decisions in uncertain environments~\cite{mnih2015human,kulkarni2016hierarchical,van2016deep,lauri2022partially}. In recent years, the emergence of deep reinforcement learning (DRL)~\cite{van2016deep} has made solving complex high-dimensional MDP problems possible. MDP based on DRL has been widely applied to numerous practical problems including robotic control~\cite{kulkarni2016hierarchical,lauri2022partially}, medical decision-making~\cite{mnih2015human},  etc, demonstrating powerful modeling capabilities and potential for solving complex decision problems. 
In this paper, we are the first to formulate semi-supervised 3D action recognition via active learning as a Markov Decision Process (MDP), and we train an informative sample selection model within the MDP framework to intelligently select samples for annotation.

\section{Method}
Given an unlabeled set of skeleton sequences and a limited annotation budget, the goal of the Semi-supervised 3D Action Recognition via AL (S3ARAL) task is to iteratively select the informative samples for annotation to maximize the performance of the target action recognizer.

Motivated by the similarity in objectives between S3ARAL and the Markov Decision Process (MDP)—where both aim to maximize long-term returns, with S3ARAL focusing on improving the action recognizer's performance throughout the active learning process and MDP targeting the maximization of expected cumulative rewards—we propose a novel perspective by formulating S3ARAL as an MDP (Section \ref{subsec:S3ARAL-MDP}).
 Specifically, we design a informative sample selection model that selects informative yet diverse skeleton sequences for annotation (Section \ref{sec:ISSM}). Moreover, to enable quick adaptation to the enlarged labeled skeleton set during deployment, we further introduce a meta tuning strategy based on meta-learning~\cite{finn2017model} (Section \ref{subsec:meta-tuning}). We introduce the overall training and testing of our method in~Section \ref{sec:training&testing}. The pipeline of our method is illustrated in Figure~\ref{fig:pipeline}.

\subsection{Task Settings and Notation}

To improve clarity, in this subsection we present the task setting and introduce the notations of the key symbols used in our method.

\noindent \textbf{Task Settings.}
In this paper, we address the task of Semi-supervised 3D Action Recognition via Active Learning (S3ARAL). 
S3ARAL operates on datasets composed of multi-dimensional time series that capture the coordinates of body keypoints over time. We denote the dataset as $X = \{ X^U \cup X^L \}$, where $X^U$ and $X^L$ represent the unlabeled and labeled subsets, respectively. Each sample $x_i \in X$ is a sequence $x_i = \left[ x_1, x_2, \ldots, x_T \right]$, where $x_t \in \mathbb{R}^{p \times d}$ denotes the coordinates of $p$ keypoints at time $t$, and $d$ is the dimensionality of the keypoints (typically $d = 3$). The number of keypoints $p$ may vary across different datasets.

Specifically, in the S3ARAL task, all samples are initially unlabeled, i.e., $X = X^U$.
With an annotation budget $B$, at the $t$-th iteration, given an unlabeled set $X_t^U$, a labeled set $X_t^L$, and an action recognizer $AR_t$, the S3ARAL method proceeds as follows:
(1) Assess the informativeness of the unlabeled samples;
(2) Select a batch of informative skeleton sequences for annotation;
(3) Transfer the selected samples from $X_t^U$ to $X_t^L$, and retrain the action recognizer $AR_t$ on the updated labeled set $X_{t+1}^L$ to obtain a new model $AR_{t+1}$.

\noindent \textbf{Notation.} Next, we introduce the meanings of some key notations used in our paper.

\textit{Skeleton Sequence.} $
x = [x_1, x_2, \dots, x_T],$
where $T$ is the sequence length and $x_t \in \mathbb{R}^{p \times d}$ denotes the 3D coordinates of $p$ joints at time step $t$, with $d=3$.

\textit{Dataset.} $X = \{ X^U \cup X^L \}$, composed of labeled and unlabeled subsets.

\textit{Budget.} $B$, the maximum number of skeleton sequences that can be annotated.

\textit{State.}
$[\widetilde{\operatorname{MMD}}_t(S^L, S^U), B^C_t]$
where $\widetilde{\mathrm{MMD}}_t$ is the hyperbolic projection of the distribution gap and $B^C_t$ is the budget consumption ratio.

\textit{Action.} $a_t$, selection of a sample based on its state.

\textit{Reward.} $r_{t+1}$, the accuracy gain of the action recognizer between iterations, measured on a reward set $X^{rwd}$.

\textit{Transition.} $(s_t, a_t, r_{t+1}, s_{t+1})$, which updates the labeled and unlabeled sets.

\subsection{Semi-supervised 3D Action Recognition via Active Learning  as a Markov Decision Process}
\label{subsec:S3ARAL-MDP}

To solve the challenge Semi-supervised 3D Action Recognition via Active Learning (S3ARAL) task, previous S3ARAL method~\cite{li2023sar} encodes skeleton sequences into a high-dimensional latent space and group the resulting latent representations into clusters. Informative samples are then selected for annotation based on their corresponding uncertainty scores and their distances within the latent space clusters in the feature space.
However, representative skeleton sequences may not always be the most informative for the action recognizer, as the model may have already learned similar patterns from previously labeled samples. Hence, selecting such redundant samples may provide limited benefit and result in sub-optimal recognition performance.
To address this, inspired by \cite{Gong2022MetaAT}, we approach S3ARAL from a novel perspective by formulating it as a Markov Decision Process (MDP), motivated by their shared objective of maximizing future returns. Besides, we treat the performance improvement of the action recognizer as the reward in our MDP.
 Such formulation enables a more intelligent  sequence sample selection strategy.

Below, we first provide  an explanation of how to view the S3ARAL task from the MDP perspective.

\noindent \textbf{New perspective on S3ARAL task.} In this paper, we approach S3ARAL from the novel MDP perspective to enable more intelligent sample selection for annotation.  Specifically, the S3ARAL process is formulated as a MDP tuple $(s_t, a_t, r_{t+1}, s_{t+1})$, with the core steps reinterpreted as follows: (1) State Estimation: Estimate the state $s_t$, which captures the distributional discrepancy between the unlabeled set $X_t^U$ and the labeled set $X_t^L$ at iteration $t$. (2) Action Selection: Evaluate each state-action pair $(s_t, a_t)$ to identify the informative skeleton sequence for annotation. (3) Environment Transition: Update $X_t^L$ and $X_t^U$ to $X_{t+1}^L$ and $X_{t+1}^U$ by moving the selected skeleton sequence from $X_t^U$ to $X_t^L$. Retrain the action recognizer $AR_t$ on the updated labeled set $X_{t+1}^L$ to obtain a new model $AR_{t+1}$, and update the state to $s_{t+1}$. (4) Reward Computation: Compute the reward $r_{t+1}$ based on the performance improvement of $AR_{t+1}$ over $AR_t$, evaluated on a separate reward set $X^{rwd}$, to guide the agent’s update.

In this way, we reformulate S3ARAL task as a MDP, allowing us to leverage the MDP framework with solid theoretical foundations~\cite{van2016deep} to perform the S3ARAL task in a more intelligent manner.

\subsection{Informative Sample Selection Model for Semi-supervised 3D Action Recognition via Active Learning }
\label{sec:ISSM}

In the previous section, we introduce how the Semi-supervised 3D Action Recognition via Active Learning (S3ARAL) task can be viewed from the perspective of the Markov Decision Process (MDP). In this section, we present the details of our proposed informative sample selection model, which enables us to perform the S3ARAL task from the MDP perspective.

In this work, we adopt the widely used $Q$-learning algorithm~\cite{Hasselt2015DeepRL} as our MDP framework, where the designed informative sample selection model (ISSM), i.e., the agent in the MDP, performs batch sampling to select skeleton samples for annotation. Specifically, our ISSM evaluates each state-action pair $(s_t, a_t)$ and selects the action $a_t$ with the highest associated $Q$-value. By deriving the reward directly from the improvement of the action recognizer, ISSM is trained to learn a policy that maximizes both the cumulative reward and the performance of the action recognizer. 
Besides, to support ISSM in making informed decisions, we design a state-action representation based on hyperbolic space~\cite{ganea2018hyperbolic}. This hyperbolic state design provides exponential volume growth, making it well-suited for representing the tree-like structure of human skeletons in a compact and expressive manner \cite{qu2024lmc}.

Below, we sequentially introduce the design of the state, action, and reward, along with the training strategy within our MDP framework.

\noindent \textbf{State.}
Intuitively, the state $s_t$ should capture the distribution gap between the labeled dataset $X_t^L$ and the unlabeled dataset $X_t^U$, enabling the agent to identify the informative skeleton sequence that can mitigate the distribution shift between $X_t^L$ and $X_t^U$. The purpose of computing this distribution gap is to make the training set more unbiased, thereby increasing the likelihood that the trained action recognizer generalizes well to unseen cases. Besides, skeleton data naturally form tree-like graphs, where hierarchical relationships exist among joints~\cite{shahroudy2016ntu}. Therefore, computing features in Euclidean space may fail to capture these structural properties effectively. In contrast, hyperbolic space exhibits exponential volume growth, allowing for efficient representation of hierarchical  skeleton data.
 Hence, to enhance the representation of the factors included in our state and action space, we transform the computed distribution gap from the Euclidean space to hyperbolic space.

Below, we first introduce how to compute the distribution gap between the labeled and unlabeled datasets, followed by the definition of hyperbolic space and the projection of the computed distribution gap from the Euclidean space to hyperbolic space.

\noindent \textbf{i) Distribution Gap.} To model the distributional drift between the labeled dataset $X_t^L$ and the unlabeled dataset $X_t^U$, we treat them as two separate domains and measure the domain gap accordingly. Specifically, we adopt the Maximum Mean Discrepancy (MMD)  to quantify the discrepancy,
as follows:
\begin{equation}
\begin{aligned}
\operatorname{MMD}(S^L, S^U) = \sum_{i=1}^{n_L} \sum_{j=1}^{n_L} \frac{k(p_i, p_j)}{n_L^2} +  \sum_{i=1}^{n_U} \sum_{j=1}^{n_U} \frac{k(q_i, q_j)}{n_U^2} \\  - \sum_{i=1}^{n_L} \sum_{j=1}^{n_U} \frac{2 \cdot k(p_i, q_j)}{n_L n_U},    
\end{aligned}
\end{equation}
where $S^L$ and $S^U$ denote the feature space distributions of $X_t^L$ and $X_t^U$, respectively,
 and the computed $\operatorname{MMD}(S^L, S^U)$ is a scalar representing the discrepancy between them. The samples from $S^L$ and $S^U$ are denoted as $p$ and $q$, respectively. $n_L$ and $n_U$ are the numbers of samples in $X_t^L$ and $X_t^U$. $k(\cdot)$ is a radial basis kernel used to compute pairwise distances.

\noindent \textbf{ii) Hyperbolic Space.} Here, inspired by \cite{qu2024llms}, we adopt the Poincaré model~\cite{ganea2018hyperbolic} as the hyperbolic space. Specifically, an $n$-dimensional Poincaré model, denoted as $\mathbb{B}^{n}_{\kappa}$, is a Riemannian manifold $(\mathbb{B}^{n}_{\kappa}, g^{\mathbb{B}}_{\mathbf{x}})$ with constant negative curvature $\kappa < 0$. The Poincaré model is defined as:
\begin{equation}
\mathbb{B}^{n}_{\kappa} = \left\{ \mathbf{x} \in \mathbb{R}^n : ||\mathbf{x}|| < -\frac{1}{\kappa}\right\},
\end{equation}
where $||\cdot||$ denotes the Euclidean norm. The manifold is further equipped with the Riemannian metric tensor $g_{\mathbf{x}}^{\mathbb{B}}$:
\begin{equation}
g^{\mathbb{B}}_{\mathbf{x}} = \left( \frac{2}{1 + \kappa ||\mathbf{x}||^2} \right)^2 g^{\mathbb{E}},
\end{equation}
where $\mathbf{x} \in \mathbb{B}^{n}_{\kappa}$ and $g^{\mathbb{E}}$ denotes the Euclidean metric tensor. This formulation highlights the strength of hyperbolic space in modeling hierarchical and structured data, where distances grow exponentially, similar to a tree structure. Such a property makes it particularly suitable for skeleton-based human action recognition, as human motion inherently follows a multi-scale hierarchical pattern.

\noindent \textbf{iii) Projection to Hyperbolic Space.} 
After introducing  definition of the hyperbolic space, which is more suitable for modeling human skeletons, we then present how to project the distribution gap computed in Euclidean space into the hyperbolic space.
Specifically, given the distribution gap $\operatorname{MMD}(S^L, S^U)$ computed in Euclidean space, we follow~\cite{ganea2018hyperbolic} to map it into hyperbolic space through the exponential map $\mathrm{exp}^c_\textbf{0}(\cdot)$ as:
\begin{equation}
\begin{aligned}
&\widetilde{\operatorname{MMD}}(S^L, S^U) = \mathrm{exp}^c_\textbf{0}\left(\operatorname{MMD}(S^L, S^U)\right) \\ &= tanh(\sqrt{c} \| \operatorname{MMD}(S^L, S^U)\|)\frac{\operatorname{MMD}(S^L, S^U)}{\sqrt{c}\|\operatorname{MMD}(S^L, S^U)\|}.
\end{aligned}
\end{equation}
Here, $\widetilde{\operatorname{MMD}}(S^L, S^U)$ in the above equation denotes the projected distribution gap $\operatorname{MMD}(S^L, S^U)$.

Moreover, the available budget in our MDP framework is a critical factor for our ISSM to make effective selections. To capture this, we incorporate the budget consumption ratio $B^C$, which is an indicator of the remaining annotation resources, into the state representation.
Hence, our state $s_t$ is defined as $[\widetilde{\operatorname{MMD}}_t(S^L, S^U), B^C_t]$, capturing both the distribution gap between the labeled and unlabeled sets and the current budget status. This state representation guides the ISSM in selecting the types of human skeletons that are beneficial for improving the performance of the action recognizer.

\noindent \textbf{Action.} The action should ideally capture the potential contribution of a specific unlabeled human skeleton when it is added to the labeled set $X_t^L$. Intuitively, by combining the state and action representations, our ISSM is expected to have sufficient information to evaluate each unlabeled human skeleton and select the informative skeleton from the unlabeled set $X_t^U$ for annotation. 
To this end, we associate each unique skeleton $x$ from the unlabeled pool $X_t^U$ with an action $a_t$ in the action space $A_t$. Specifically, to facilitate the selection of informative samples, we extract two types of features from each unlabeled skeleton sequence: (1) the skeleton sequence's potential contribution to improving the action recognizer, measured by its uncertainty. (2) the representativeness of the skeleton sequence within the unlabeled dataset. Below, we sequentially introduce the computation methods of these two types of action representations.

We adopt the widely used marginal index (MI) to measure the uncertainty~\cite{roth2006margin}. Specifically, for a skeleton sample $x$, MI quantifies the confidence gap between the most and the second most probable action classes predicted by the action recognizer, and $\text{MI}(x)$ is calculated as:
\begin{equation}
\text{MI}(x)=1-\left(\max_C p(c \mid x)-\max_{C \neq c^*}p(c \mid x)\right),
\end{equation}
where 
\begin{equation}
c^*=\arg \max _Cp(c\mid x).
\end{equation}
Here, $C$ denotes all action classes, $c^*$ denotes the action class with the highest predicted probability by the action recognizer, and $\displaystyle \max_{\substack{c \neq c^*}}p(c\mid x)$ denotes the second highest predicted probability.

We then present our parameterization of the skeleton sequence representativeness. 
Specifically, we select representative samples with respect to the unlabeled set $X_t^U$, by leveraging the distribution of similarity scores,  and introduce a histogram-based representation $R$ that encodes the cosine similarity distribution between a sample $x$ and the average behavior of samples in $X_t^U$ across the feature space of the action recognizer. 
Including such a factor in our action space allows the ISSM to avoid repeatedly selecting representative action sequences that have already been learned by the action recognizer, thereby improving sampling efficiency.

It is worth noting that for the features in the action space, we follow the approach used in the state representation and project them from Euclidean space to hyperbolic space.

\noindent \textbf{Reward.} The reward serves as a metric to quantify the benefit that a selected unlabeled skeleton sequence brings to our action recognizer $AR_{t}$ at iteration $t$. To enable accurate reward estimation, we reserve a reward subset $X^{rwd}$ prior to our active learning procedure. The reward $r_{t+1}$ is defined as the accuracy gain of the action recognizer on $X^{rwd}$, computed as the difference in performance between $AR_{t+1}$ and $AR_t$. Note that $X^{rwd}$ is used solely for evaluation and is excluded from the training of the action recognizer. With the reward signal $r_{t+1}$, our ISSM can be optimized to select the informative skeleton sequence, thereby improving 3D action recognition accuracy in the active learning iteration.

\noindent \textbf{Training of the ISSM.}  To train our ISSM, we adopt the Double DQN framework~\cite{Hasselt2015DeepRL} and optimize the model by minimizing the temporal difference (TD) error:
\begin{equation}
\begin{aligned}
TD(\theta, \hat{\theta}) = & (AR_t(s_t, a_t ; \theta) - r_{t+1}  \\
&  - \gamma \cdot  AR_{t+1}(s_{t+1}, a_{t+1}; \hat{\theta}) )^2
\end{aligned}
\end{equation}
Here, $\theta$ denotes the parameters of our ISSM, and  $\hat{\theta}$ corresponds to the parameters of the target (off-policy) model, which maintains the learned $Q$-values and is periodically updated with $\theta$, following the Double DQN setup.

Overall, the carefully designed state and action spaces of our ISSM, combined with the theoretically grounded MDP training framework, enable our ISSM to more intelligently select skeleton sequences for annotation.

\subsection{Meta Tuning}
\label{subsec:meta-tuning}

When deploying our trained informative sample selection mode (ISSM) to real-world scenarios, it typically requires re-training on an enlarged labeled dataset. This re-training process could be time-consuming due to the continuously growing dataset size. To address this issue, we design a meta tuning strategy based on meta-learning~\cite{finn2017model}, which aims to enhance the generalization ability of our ISSM and thereby accelerates its deployment.
Specifically, our proposed meta tuning formulates the re-training of our ISSM on the expanded labeled dataset as a meta-learning task, 
which consists of three
stages: virtual-train, virtual-test, and meta-update. We consider our ISSM training in source dataset, i.e., training on the unexpanded labeled dataset, as a virtual-train task, and re-trained our ISSM on the expanded labeled dataset as a virtual-test task.
Virtual-test provides feedback on how to optimize the ISSM in 
a manner that generalizes across different datasets. Subsequently, the meta-update phase consolidates the 
feedback from the first two phases and applies the actual 
parameter updates to our model, resulting in improved 
generalization capabilities that promote fast adaptation.

 To simulate the meta tuning process, we partition initial training dataset $X_{\text{i}}$ into a smaller subset as the virtual-train set $X_{\text{i}}^{vtr}$ and a larger subset as the virtual-test set $X_{\text{i}}^{vte}$. Our objective is to learn meta parameters ${\theta}_{\text{mt}}$ that can adapt effectively to $X_{\text{i}}^{vte}$ after being optimized on $X_{\text{i}}^{vtr}$. Specifically, starting from randomly initialized meta parameters ${\theta}_{\text{mt}}$, we first update our ISSM on $X_{\text{i}}^{vtr}$ to obtain ${\theta}_{\text{mt}}^*$. We then use the updated ISSM parameterized by ${\theta}_{\text{mt}}^*$ to perform active learning on $X_{\text{i}}^{vtr}$ and compute the meta loss, which is used to refine ${\theta}_{\text{mt}}^*$. The meta loss is  the Temporal Difference (TD) error:
\begin{equation}
\begin{aligned}
TD_{\text{mt}}({\theta}_{\text{mt}}^*&) = \sum_{t=0}^{H-1} (   AR_t(s_t, a_t; \theta_{\text{mt}}^*) \\& - r_{t+1} - \gamma \cdot AR_{t+1}(s_{t+1}, a_{t+1}; \theta_{\text{mt}}^*) )^2.
\end{aligned}
\end{equation}
Here, $H$ denotes the number of iterations in our meta tuning process.
After obtaining the meta loss $TD_{\text{mt}}({\theta}_{\text{mt}}^*)$, we update the meta parameters according to:
\begin{equation}
{\theta}_{\text{mt}} = {\theta}_{\text{mt}} - \beta \cdot \nabla TD_{\text{mt}}({\theta}_\text{mt}^*).
\end{equation}
Here, $\beta$ is the learning rate for meta tuning.
By minimizing the meta loss $TD_{\text{mt}}$, we obtain the meta parameters ${\theta}^*_{\text{mt}}$ that enable fast adaptation to the  virtual-test set $X_{\text{i}}^{vte}$ using updates from only the  virtual-train set $X_{\text{i}}^{vtr}$.

Overall, the meta tuning strategy enhances the generalization ability of our ISSM, thereby accelerating its deployment in real-world scenarios.

\subsection{Overall Training and Testing}
\label{sec:training&testing}

In the previous sections, we formulate Semi-supervised 3D Action Recognition via Active Learning (S3ARAL) as a Markov Decision Process (MDP), allowing us to leverage the well-established MDP framework to perform S3ARAL in a more intelligent manner (Section \ref{subsec:S3ARAL-MDP}). Then, we introduce the informative sample selection model (ISSM) for selecting skeleton data for annotation (Section \ref{sec:ISSM}), as well as a meta-tuning strategy to enable rapid adaptation of ISSM in real-world scenarios (Section \ref{subsec:meta-tuning}). In this section, we provide an overview of the training and testing procedures of our method, along with an algorithm that outlines the overall framework (see Algorithm~\ref{alg:framework}).

\noindent \textbf{Training.} Given an unlabeled dataset $X$ and an annotation budget $B$, our method proceeds as follows: We first randomly sample an initial subset $X_{\text{i}}$ for annotation. Using the labeled initial subset $X_{\text{i}}$, we then simulate the S3ARAL process by partitioning $X_{\text{i}}$ and training our ISSM. Specifically, we divide the labeled initial set $X_{\text{i}}$ into a labeled set $X_{\text{i}}^L$, an unlabeled set $X_{\text{i}}^U$, and a reward set $X^{rwd}$, and then allow our ISSM to select the  skeleton sequence  as described in Section~\ref{sec:ISSM}.
Furthermore, once our ISSM is trained on $X_{\text{i}}$, it can be deployed to perform the actual S3ARAL process on the remaining unlabeled pool $X^U = X \backslash X_{\text{i}}$, until the annotation budget $B$ is exhausted. We refer to this stage as the deployment phase, during which our ISSM selects batch skeleton data $\left\{x^n\right\}_{n=1}^N$ from $X^U$ for annotation at each iteration, gradually expanding the labeled pool $X^L = X^L \cup \left\{x^n\right\}_{n=1}^N$ to update our action recognizer $AR$. We initialize $X^L = X_{\text{i}}$ at the beginning of this phase.
Besides, to accelerate the deployment of our method in the deployment phase, we design a meta tuning strategy for our ISSM (Section \ref{subsec:meta-tuning}).

\noindent \textbf{Testing.}
After deployment, our ISSM remains frozen and automately selects samples for annotation.

\begin{algorithm}[t]
\caption{Overall Framework of the Proposed Method}
\label{alg:framework}
\begin{algorithmic}[1]
\Require Unlabeled set $X^U$, budget $B$, initial labeled set $X_0$, reward set $X^{rwd}$
\Ensure Final trained action recognizer

\State \textbf{Stage 1: Active Learning with Limited Budget}
\State Initialize $X^L \gets X_0$, $X^U \gets X^U \setminus X_0$
\While{$|X^L| < B$ and $X^U \neq \varnothing$}
   \State Construct state features 
   \State Estimate quality scores of all candidates in $X^U$
   \State Select top samples and add to $X^L$
   \State Remove selected samples from $X^U$
   \State Retrain the recognition model on $X^L$
   \State Evaluate on $X^{rwd}$ and update model
\EndWhile

\State \textbf{Stage 2: Meta Tuning}
\State Split $X_0$ into virtual training and validation subsets
\For{each meta-iteration}
   \State Update the model using virtual training set
   \State Validate on virtual validation set and refine parameters
\EndFor
\State \Return Final trained recognition model 
\end{algorithmic}
\end{algorithm}

\section{Experiments}

\subsection{Evaluation Dataset and Metric}

For a fair comparison, we follow~\cite{li2023sar} to evaluate our method on three widely used benchmark datasets: UWA3D Multiview Activity II (UWA3D)~\cite{Rahmani2014HOPCHO}, North-Western UCLA (NW-UCLA)~\cite{Wang2014CrossViewAM}, and NTU RGB+D 60~\cite{Shahroudy2016NTURA}.
These datasets vary in the number of action classes and include both cross-view  and cross-subject evaluation settings. We follow~\cite{li2023sar} and use classification accuracy as the evaluation metric. Below, we provide a detailed introduction to the three evaluation datasets used in this work.

\noindent \textbf{UWA3D} consists of 30 human action categories, with each action performed four times by ten subjects and recorded from four viewpoints: frontal, left, right, and top. Following~\cite{li2023sar}, we use the frontal and left views for training and reserve the right view for testing to ensure a fair comparison.
 This setup presents a more challenging evaluation scenario due to the increased view discrepancy.

\noindent \textbf{NW-UCLA}  is collected using Kinect v1 cameras and consists of 1,494 samples performed by 10 subjects. It includes 10 action classes, with each skeleton comprising 20 joints. Following the evaluation protocol in~\cite{li2023sar}, we use samples from camera views 1 and 2 as the training set, while samples from camera view 3 serve as the test set. 

\noindent \textbf{NTU RGB+D 60}, captured with a Kinect v2 sensor, is the largest publicly available benchmark for depth-based action recognition, containing over 56,000 video sequences and 4 million frames. It includes recordings from 80 different viewpoints and covers 60 action classes, ranging from daily activities and medical conditions to interactive actions, performed by 40 subjects aged between 10 and 35. The dataset poses considerable challenges due to large intra-class variation and diverse viewpoints. Owing to its scale, it is particularly suitable for deep learning-based activity recognition. Models pre-trained on this dataset can be effectively fine-tuned on smaller datasets, leading to faster convergence and improved performance. In the NTU RGB+D 60 dataset, each sample  provides 3D coordinates of 25 body joints, offering rich skeletal information for detailed analysis.

\subsection{Baseline Methods}
To ensure a comprehensive and fair comparison with previous Semi-supervised 3D Action Recognition via Active Learning method, we follow the settings in~\cite{li2023sar} and compare our approach with state-of-the-art (SOTA) active learning and semi-supervised methods.

In particular, we compare our method against four representative active learning approaches: uniform sampling~\cite{li2023sar}, core set selection~\cite{Sener2017ActiveLF}, discriminator-based selection~\cite{Sinha2019VariationalAA}, and consistency-based active learning under augmentation~\cite{Gao2019ConsistencyBasedSA}. Specifically, in uniform sampling, samples are randomly selected for annotation from the entire dataset. Core-set selection aims to cover the feature space by choosing samples that minimize the overall coverage radius. Discriminator-based selection employs a discriminator to distinguish whether a sample is labeled. Consistency-based active learning under augmentation selects samples based on the consistency of augmented skeleton sequences.

Besides, following~\cite{li2023sar}, we compare our method with three representative semi-supervised skeleton-based 3D action recognition methods: ASSL~\cite{si2020adversarial}, MS$^2$L~\cite{lin2020ms2l}, and SC3D~\cite{thoker2021skeleton}. It is important to note that these methods assume access to a pre-defined labeled set and do not address the problem of active sample selection.

\subsection{Implementation Details}
For fair comparison, we follow~\cite{li2023sar} and adopt an encoder-decoder architecture to encode skeleton data. Specifically, our encoder consists of three-layer bi-GRU cells with 1024 hidden units in each direction. The hidden states from both directions are concatenated into a 2048-dimensional latent representation, which is then fed into the decoder. The decoder is a uni-directional GRU with a hidden size of 2048. 
For a fair comparison, we follow~\cite{li2023sar} and use the Adam optimizer~\cite{Kingma2014AdamAM} to train the action recognizer. Both the encoder–decoder, optimized with a reconstruction loss, and the action recognizer, optimized with a classification loss, are trained simultaneously. The learning rate is initialized to $10^{-4}$ and decayed by a factor of $0.95$ every 10 epochs on the UWA3D and NW-UCLA datasets, and every 3 epochs on the NTU RGB+D 60 dataset.
Our informative sample selection model (ISSM) is a lightweight three-layer MLP. For a fair comparison, AL-SAR and several active learning baselines adopt the same three-layer MLP as the action recognizer, consistent with our method. We also use the Adam optimizer~\cite{Kingma2014AdamAM} as its optimizer. The learning rate for the ISSM is set to $1 \times 10^{-4}$  
across all three datasets: UWA3D, NW-UCLA, and NTU RGB+D 60. Specifically, our model is trained for 100 epochs on the UWA3D and NW-UCLA datasets, and for 300 epochs on the NTU RGB+D dataset. Our ISSM is trained for 50 epochs on the UWA3D and NW-UCLA datasets, and for 100 epochs on the NTU RGB+D 60 dataset. The model is pre-trained in advance and kept frozen during our Semi-supervised 3D Action Recognition via Active Learning task.

\begin{table*}[htbp]
\caption{Quantitative comparisons  with representative active learning methods (Uniform sampling~\cite{li2023sar}, Core Set selection~\cite{Sener2017ActiveLF}, discriminator-based selection~\cite{Sinha2019VariationalAA}, and consistency-based active learning under augmentation~\cite{Gao2019ConsistencyBasedSA}, and AL-SAR~\cite{li2023sar}) on three common 3D action recognition datasets (UWA3D~\cite{Rahmani2014HOPCHO}, NW-UCLA~\cite{Wang2014CrossViewAM}, and NTU RGB+D 60~\cite{Shahroudy2016NTURA}) under different proportions of labeled samples.
``\%Labels'' denotes the proportion of labeled samples relative to the total number of training samples in the dataset, while ``\#Labels'' indicates the absolute number of labeled samples.  ``VIEW3'' refers to using the third camera view of the multi-view UWA3D dataset as the testing set, while the remaining views are used for training. ``CS'' stands for ``Cross-Subject'', meaning that the dataset is split by subjects, where the actions of some subjects are used for training and those of the remaining subjects are used for testing. The results of ``Ours'' are obtained by running our method five times. \colorbox{Best}{Best} and \colorbox{Second}{Second Best} results are highlighted.}
\label{tab:main-results}
\centering
\resizebox{\linewidth}{!}{
\begin{tabular}{c|l|cccc|cccc|cccc}
 \toprule[1.2pt] & Datasets & \multicolumn{4}{c}{ UWA3D VIEW3~\cite{Rahmani2014HOPCHO} } & \multicolumn{4}{|c|}{ NW-UCLA~\cite{Wang2014CrossViewAM} } & \multicolumn{4}{c}{ NTU RGB+D 60 CS~\cite{Shahroudy2016NTURA} } \\
\midrule & \%Labels & 5\% & 10\% & 20\% & 50\% & 5\% & 15\% & 30\% & 40\% & 1\% & 2\% & 5\% & 10\% \\
& \#Labels
 & 25 & 50 & 100 & 250 & 50 & 150 & 300 & 400 & 400 & 800 & 2000 & 4000 \\
\midrule \multirow{5}{*}{\rotatebox[origin=c]{90}{Baselines}}  & Discriminator-based Selection~\cite{Sinha2019VariationalAA} & 19.3 & 28.7 & 40.2 & 53.8 & 47.7 & 71.8 & 76.6 & 80.5 & 34.9 & 39.5 & 53.8 & 60.4 \\
 & Core Set~\cite{Sener2017ActiveLF} & 21.5 & 29.9 & 40.3 & 52.6 & 57.3 & 69.4 & 77.3 & 80.6 & 17.6 & 23.1 & 37.0 & 49.6 \\
 & Consistency-based AL under Augmentation~\cite{Gao2019ConsistencyBasedSA} & 21.9 & 30.2 & 40.8 &\cellcolor{Second}56.2 & 48.2 & 65.8 & 77.0 & 81.7 & 20.7 & 33.5 & 50.4 & 60.2 \\
  & Uniform Sampling~\cite{li2023sar} & 23.4 & 30.5 & 42.1 & 53.4 & 52.3 & 70.4 & 78.4 & 80.8 & 34.6 & 43.8 & 56.3 & 61.2 \\
 & AL-SAR~\cite{li2023sar} & \cellcolor{Second}25.7 & \cellcolor{Second}35.3 & \cellcolor{Second}45.7 & 53.9 & \cellcolor{Second}61.0 & \cellcolor{Second}75.9 & \cellcolor{Second}82.5 & \cellcolor{Second}84.1 & \cellcolor{Second}38.8 & \cellcolor{Second}47.6 & \cellcolor{Second}57.9 & \cellcolor{Second}63.7 \\
 \midrule
 & Ours & \cellcolor{Best}28.9 $\pm$ 0.2  & \cellcolor{Best}39.1 $\pm$ 0.1 & \cellcolor{Best}49.1 $\pm$ 0.3 & \cellcolor{Best}58.7 $\pm$ 0.1 & \cellcolor{Best}64.3 $\pm$ 0.2 & \cellcolor{Best}79.1 $\pm$ 0.1 & \cellcolor{Best}86.0 $\pm$ 0.2 & \cellcolor{Best}87.9 $\pm$ 0.2 & \cellcolor{Best}41.1 $\pm$ 0.3 & \cellcolor{Best}51.3 $\pm$ 0.1 & \cellcolor{Best}60.2 $\pm$ 0.3 & \cellcolor{Best}66.9 $\pm$ 0.2 \\
\bottomrule[1.2pt]
 \end{tabular}
 }
\end{table*}

\subsection{Main Experiments}
In the previous subsection, we introduce the evaluation datasets and metric, the baselines for comparison, and the implementation details of our method. In this subsection, we present the comparison results of our method against state-of-the-art (SOTA) active learning baselines on the three datasets, i.e., UWA3D, NW-UCLA, and NTU RGB+D 60.

\begin{table}[htbp]
\caption{We compare the performance of action recognizers trained on samples obtained through random selection, as used in previous semi-supervised action recognition methods (MS$^2$L~\cite{lin2020ms2l}, and SC3D~\cite{thoker2021skeleton}), with those trained on samples selected by our method and AL-SAR~\cite{li2023sar},
on the widely-used 3D action recognition datasets NTU RGB+D 60~\cite{Shahroudy2016NTURA} under different proportions of labeled samples.
``\%Labels'' denotes the proportion of labeled samples relative to the total number of training samples in the dataset, while ``\#Labels'' indicates the absolute number of labeled samples.  \colorbox{Best}{Best} results are highlighted.}
\label{tab:main-results-ss}
\centering
\resizebox{0.42\textwidth}{!}{
\begin{tabular}{l|cccc}
 \toprule[1.2pt]  Datasets   & \multicolumn{4}{c}{ NTU RGB+D 60 CS~\cite{Shahroudy2016NTURA} } \\
\midrule  \%Labels   & 1\% & 2\% & 5\% & 10\% \\
 \#Labels
  & 400 & 800 & 2000 & 4000 \\
\midrule  
 MS$^2$L~\cite{lin2020ms2l}  & 33.1 & 40.3 & 57.8 & 65.2 \\
 MS$^2$L~\cite{lin2020ms2l} + AL-SAR~\cite{li2023sar}  & 35.3 & 42.8 & 59.7 & 67.0 \\
 MS$^2$L~\cite{lin2020ms2l} + Ours  & \cellcolor{Best}38.0 & \cellcolor{Best}45.9 & \cellcolor{Best}61.2 & \cellcolor{Best}69.8 \\
 \midrule 
 SC3D~\cite{thoker2021skeleton}  & 35.7 & 41.2 & 59.6 & 65.9 \\
 SC3D~\cite{thoker2021skeleton} + AL-SAR~\cite{li2023sar}  & 36.6 & 42.8 & 61.0 & 67.3 \\
  SC3D~\cite{thoker2021skeleton} + Ours  & \cellcolor{Best}39.2 & \cellcolor{Best}45.6 & \cellcolor{Best}63.8 & \cellcolor{Best}69.5 \\
\bottomrule[1.2pt]
 \end{tabular}
 }
\end{table}

\noindent \textbf{Comparison With SOTA Active Learning Methods.} As shown in Table~\ref{tab:main-results}, we compare our method with previous state-of-the-art active learning approaches on commonly used 3D action recognition datasets under different numbers of labeled samples. As can be seen in Table~\ref{tab:main-results}, our method significantly outperforms previous SOTA active learning methods. For example, on the UWA3D dataset with 50 labeled samples, our method achieves an accuracy of 39.1\%, surpassing the previous best method AL-SAR by 3.8\%.

Our method also demonstrates strong robustness across datasets of different scales, including the relatively smaller UWA3D and NW-UCLA datasets, as well as the larger NTU RGB+D 60 dataset.
 For example, on the large-scale NTU RGB+D 60 dataset with 800 labeled samples, our method achieves an accuracy of 51.3\%, outperforming the previous SOTA method AL-SAR, which achieves 47.6\%, by 3.7\%. In addition, on the moderately sized NW-UCLA dataset, our method consistently outperforms the prior SOTA method. For instance, under the setting of 400 labeled samples, our method achieves 87.9\% accuracy, while AL-SAR only reaches 84.1\%, resulting in a 3.8\% improvement. Moreover, as shown in Table~\ref{tab:main-results}, our method also exhibits strong robustness under varying numbers of labeled samples within the same dataset. For example, on the NW-UCLA dataset, our method achieves 64.3\% accuracy with only 50 labeled samples, and maintains high performance with more labeled samples, reaching 87.9\% accuracy with 400 samples.

In addition, we combine our method and AL-SAR with two state-of-the-art semi-supervised skeleton-based action recognition methods, MS$^2$L and SC3D, to select training samples, and evaluate their performance on the NTU RGB+D 60 dataset (see Table~\ref{tab:main-results-ss}). Experimental results show that MS$^2$L and SC3D achieve better action recognition performance when trained on samples selected by our method. For example, under the 4,000 labeled samples setting on the NTU RGB+D 60 dataset, MS$^2$L + AL-SAR achieves an accuracy of 67.0\%, while MS$^2$L + Ours achieves 69.8\%.

Overall, our method demonstrates strong robustness across different datasets and labeling ratios, consistently outperforming previous SOTA methods. We attribute this to the use of a theoretically grounded MDP framework for training our informative sample selection model, enabling more intelligent sample selection.

\begin{table*}[htbp]
\caption{Ablation study on our designed state (Distribution Gap) and action representations (Marginal Index and Skeleton Sequence Representativeness) across three commonly used 3D action recognition datasets (UWA3D~\cite{Rahmani2014HOPCHO}, NW-UCLA~\cite{Wang2014CrossViewAM}, and NTU RGB+D 60~\cite{Shahroudy2016NTURA}) under varying proportions of labeled samples.
``\%Labels'' denotes the proportion of labeled samples relative to the total number of training samples in the dataset, while ``\#Labels'' indicates the absolute number of labeled samples. \colorbox{Best}{Best} results are highlighted in the table.} 
\label{tab:ablation-state}
\centering
\begin{tabular}{l|l|cc|cc|cc}
 \toprule[1.2pt] Type &Datasets & \multicolumn{2}{c}{ UWA3D VIEW3~\cite{Rahmani2014HOPCHO} } & \multicolumn{2}{|c|}{ NW-UCLA~\cite{Wang2014CrossViewAM} } & \multicolumn{2}{c}{ NTU RGB+D 60 CS~\cite{Shahroudy2016NTURA}}  \\
\midrule & \%Labels & 5\%  & 20\% & 5\%   & 30\%  & 1\%  & 5\%  \\
& \#Labels
 & 25  & 100  & 50  & 300  & 400  & 2000  \\
  \midrule
\multirow{2}{*}{State} & Ours w/o Distribution Gap &18.1  & 38.7  & 45.6  & 73.1  & 15.9  & 38.1  \\
 & Ours w/o Budget Consumption Ratio $B^C$ &25.1  & 45.5  & 52.8  & 79.0  & 36.3  & 54.4  \\
  \midrule
\multirow{2}{*}{Action} & Ours w/o Marginal Index &24.1  & 45.3  & 52.1  & 78.7  & 35.6  & 54.2  \\
 & Ours w/o Skeleton Sequence Representativeness &25.6  & 46.0  & 53.9  & 79.1  & 37.0  & 54.9  \\
 \midrule
  & Ours &\cellcolor{Best}28.9  & \cellcolor{Best}49.1  & \cellcolor{Best}64.3  & \cellcolor{Best}86.0  & \cellcolor{Best}41.1  & \cellcolor{Best}60.2  \\
\bottomrule[1.2pt]
 \end{tabular}
\end{table*}

\begin{table*}[htbp]
\caption{Ablation Study on Our Designed Hyperbolic Representation on three common 3D action recognition datasets (UWA3D~\cite{Rahmani2014HOPCHO}, NW-UCLA~\cite{Wang2014CrossViewAM}, and NTU RGB+D 60~\cite{Shahroudy2016NTURA}) under different proportions of labeled samples.
``\%Labels'' denotes the proportion of labeled samples relative to the total number of training samples in the dataset, while ``\#Labels'' indicates the absolute number of labeled samples. \colorbox{Best}{Best} results are highlighted.}
\label{tab:ablation-Hyperbolic}
\centering
\begin{tabular}{l|cc|cc|cc}
 \toprule[1.2pt]  Datasets & \multicolumn{2}{c}{ UWA3D VIEW3~\cite{Rahmani2014HOPCHO} } & \multicolumn{2}{|c|}{ NW-UCLA~\cite{Wang2014CrossViewAM} } & \multicolumn{2}{c}{ NTU RGB+D 60 CS~\cite{Shahroudy2016NTURA} } \\
\midrule  \%Labels & 5\%  & 20\% & 5\%   & 30\%  & 1\%  & 5\%  \\
 \#Labels
 & 25  & 100  & 50  & 300  & 400  & 2000  \\
  \midrule
   Ours w/o Hyperbolic Representation &26.8  & 47.1  & 60.5  & 82.3  & 39.0  & 57.8  \\
  \midrule
   Ours &\cellcolor{Best}28.9  & \cellcolor{Best}49.1  & \cellcolor{Best}64.3  & \cellcolor{Best}86.0  & \cellcolor{Best}41.1  & \cellcolor{Best}60.2  \\
\bottomrule[1.2pt]
 \end{tabular}
\end{table*}

\subsection{Ablation Studies}
In the previous subsection, we present the main experiments of our work, comparing our method with previous SOTA active learning methods on three widely used skeleton-based action recognition datasets.
In this subsection, we first follow AL-SAR~\cite{li2023sar} to verify the performance gains brought by the active learning framework in our method. In addition, we analyze the impact of our state and action space designs, the projection of representations from Euclidean space to hyperbolic space, and the effectiveness of the meta tuning strategy. Finally, we analysis the generalization of our method.

We follow AL-SAR~\cite{li2023sar} to conduct our ablation studies under the settings of 25 and 100 labeled samples on the UWA3D dataset, 50 and 300 labeled samples on the NW-UCLA dataset, and 400 and 2000 labeled samples on the NTU RGB+D 60 dataset.

\begin{table*}[htbp]
\caption{Ablation Study of our designed meta tuning on three common 3D action recognition datasets (UWA3D~\cite{Rahmani2014HOPCHO}, NW-UCLA~\cite{Wang2014CrossViewAM}, and NTU RGB+D 60~\cite{Shahroudy2016NTURA}) under different proportions of labeled samples. ``Time'' denotes the duration required for the model to converge from the start of training.
``\#Labels'' indicates the absolute number of labeled samples. \ding{51} and \ding{55} represent whether meta-tuning is performed or not, respectively.} 
\label{tab:ablation-meta}
\centering
\begin{tabular}{llccccc}
 \toprule[1.2pt] Datasets &\#Labels & Meta Tuning & Accuracy (\%) & Time (hour)  \\
\midrule \multirow{4}{*}{UWA3D VIEW3~\cite{Rahmani2014HOPCHO}} & 25 & \ding{51}  & 28.8   & 0.4    \\
 & 25 & \ding{55}   & 28.9   & 0.7    \\
& 100
 &\ding{51}   & 48.7   & 1.0    \\
& 100
 & \ding{55}    & 49.1   & 2.5    \\
  \midrule \multirow{4}{*}{NW-UCLA~\cite{Wang2014CrossViewAM}} & 50 &\ding{51}  & 64.2   & 0.4   \\
  & 50 & \ding{55}   & 64.3   & 0.8   \\
& 300
 &\ding{51}  & 85.8  & 1.7    \\
& 300
 & \ding{55}    & 86.0  & 3.3    \\
\midrule \multirow{4}{*}{NTU RGB+D 60 CS~\cite{Shahroudy2016NTURA} } & 400 &\ding{51}  & 40.9   & 0.6    \\
 & 400 & \ding{55}   & 41.1   & 1.0   \\
& 2000
 &\ding{51}   & 59.9  & 2.5    \\
& 2000
 & \ding{55}    & 60.2  & 5.1    \\
\bottomrule[1.2pt]
 \end{tabular}
\end{table*}

\noindent \textbf{Impact of State and Action Space.} As shown in Table~\ref{tab:ablation-state}, we conduct ablation studies on the distribution gap in the state and the marginal index and skeleton sequence representativeness in the action across three commonly used skeleton-based action recognition datasets. In the setting ``Ours w/o Distribution Gap'', the state is randomly generated. The results show that removing the Distribution Gap from the state leads to a 10.8\% drop in accuracy, demonstrating its effectiveness in helping our method identify informative skeleton sequences and train a better action recognizer. From the experimental results of ``Ours w/o Budget Consumption Ratio $B^C$'', we observe that incorporating $B^C$ into the state leads to further performance improvements. We attribute this to the ability of $B^C$ to help the method better recognize the current learning stage of the model, which in turn facilitates the selection of samples that are more suitable for the corresponding training stage.

Furthermore, the ablation results on the action space confirm the effectiveness of both the Marginal Index and Skeleton Sequence Representativeness. These two factors are shown to be complementary and jointly contribute to the performance improvement of our method.

\noindent \textbf{Impact of Hyperbolic Representation.} To better classify human actions based on skeletal data, we design a Hyperbolic Representation tailored to the tree-like structure of human skeletons for both state and action representations. Here, we conduct an ablation study on this Hyperbolic Representation. As shown in Table~\ref{tab:ablation-Hyperbolic}, applying the Hyperbolic Representation leads to improved performance, demonstrating the effectiveness of this representation space.

\noindent \textbf{Impact of Meta Tuning.}
To accelerate the deployment of our method in real-world scenarios, we introduce a meta tuning strategy based on meta-learning~\cite{finn2017model}. Here, we conduct an ablation study to evaluate its effectiveness (see Table~\ref{tab:ablation-meta}). In the table, "Time" denotes the convergence time required for training. As shown by the results, our meta tuning significantly speeds up model convergence. For example, under the NTU RGB+D 60 dataset with 2000 labeled samples, ``\ding{51}'' converges in only 2.5 hours, compared to 5.1 hours for ``\ding{55}'', achieving nearly a 50\% reduction in training time with comparable accuracy. Moreover, across all three datasets, ``\ding{51}'' consistently achieves similar accuracy to ``\ding{55}'' while reducing convergence time by roughly half, demonstrating the robustness of the proposed meta tuning strategy.

\begin{table}[t]
\caption{Quantitative comparisons  with representative active learning methods (Uniform sampling~\cite{li2023sar}, Core Set selection~\cite{Sener2017ActiveLF}, discriminator-based selection~\cite{Sinha2019VariationalAA}, and consistency-based active learning under augmentation~\cite{Gao2019ConsistencyBasedSA}, and AL-SAR~\cite{li2023sar}) on three common 3D action recognition datasets (UWA3D~\cite{Rahmani2014HOPCHO}, NW-UCLA~\cite{Wang2014CrossViewAM}, and NTU RGB+D 60~\cite{Shahroudy2016NTURA}) under different proportions of labeled samples.
``\%Labels'' denotes the proportion of labeled samples relative to the total number of training samples in the dataset, while ``\#Labels'' indicates the absolute number of labeled samples.  ``VIEW3'' refers to using the third camera view of the multi-view UWA3D dataset as the testing set, while the remaining views are used for training. ``CS'' stands for ``Cross-Subject'', meaning that the dataset is split by subjects, where the actions of some subjects are used for training and those of the remaining subjects are used for testing. The results of ``Ours'' are obtained by running our method five times.} 
\label{tab:generalization}
\centering
\resizebox{0.47\textwidth}{!}{
\begin{tabular}{c|l|cccc}
 \toprule[1.2pt] & Datasets  & \multicolumn{4}{c}{ NW-UCLA~\cite{Wang2014CrossViewAM} }  \\
\midrule & \%Labels & 5\% & 15\% & 30\% & 40\% \\
& \#Labels
  & 50 & 150 & 300 & 400  \\
\midrule \multirow{5}{*}{\rotatebox[origin=c]{90}{Baselines}}  & Discriminator-based Selection~\cite{Sinha2019VariationalAA}  & 47.7 & 71.8 & 76.6 & 80.5  \\
 & Core Set~\cite{Sener2017ActiveLF}  & 57.3 & 69.4 & 77.3 & 80.6  \\
 & Consistency-based AL under Augmentation~\cite{Gao2019ConsistencyBasedSA}  & 48.2 & 65.8 & 77.0 & 81.7 \\
  & Uniform Sampling~\cite{li2023sar}  & 52.3 & 70.4 & 78.4 & 80.8 \\
 & AL-SAR~\cite{li2023sar}  & 61.0 &75.9 & 82.5 & 84.1 \\
 \midrule
   & Ours w/o retraining  & \cellcolor{Second}63.2 & \cellcolor{Second}77.8  & \cellcolor{Second}85.3  & \cellcolor{Second}86.5   \\
 & Ours w/ retraining  & \cellcolor{Best}64.3 & \cellcolor{Best}79.1  & \cellcolor{Best}86.0  & \cellcolor{Best}87.9   \\

\bottomrule[1.2pt]
 \end{tabular}
 }
\end{table}

\noindent \textbf{Generalization Ability.}
To demonstrate the generalization capability of our method, we add new experiments. Specifically, the newly added experiments evaluate the generalization of our Informative Sample Selection Model by training it with 100 samples on the UWA3D VIEW3 dataset and testing it on the NW-UCLA dataset. As shown in the experimental results (see \ref{tab:generalization}), our method surpasses previous approaches even without being trained on the NW-UCLA dataset, demonstrating its strong generalization ability.

\section{Conclusion}

In this paper, we propose a novel perspective for the semi-supervised 3D action recognition via active learning task by formulating it as a Markov Decision Process (MDP), aiming to address the limitation of previous margin-based selection strategy that may fail to select samples that effectively enhance the performance of the action recognizer. 
Specifically, we leverage the theoretically grounded MDP framework to train an informative sample selection model that intelligently selects representative samples for annotation. To enhance the representational capacity of the state-action pairs in our MDP framework, we map them from the Euclidean space to hyperbolic space. Moreover, we explore a meta tuning strategy based on meta-learning to accelerate the deployment of our method in real-world scenarios. Extensive experiments are conducted on three widely used 3D action recognition datasets: UWA3D, North-Western UCLA, and NTU RGB+D 60. The results show that our method yields greater improvements in action recognizer performance compared to prior approaches and exhibits strong generalization capability.

\section*{Acknowledgments}
This work was supported by the National Key Research and Development Program of China No. 2024YFC3015600, the Fundamental Research Funds for Central Universities No.2042023KF0180 \& No.2042025KF0053. The numerical calculation is supported by super-computing system in Super-computing Center of Wuhan University.

\bibliographystyle{IEEEtran}
\bibliography{IEEEabrv,main}

\begin{thebibliography}{10}
\providecommand{\url}[1]{#1}
\csname url@samestyle\endcsname
\providecommand{\newblock}{\relax}
\providecommand{\bibinfo}[2]{#2}
\providecommand{\BIBentrySTDinterwordspacing}{\spaceskip=0pt\relax}
\providecommand{\BIBentryALTinterwordstretchfactor}{4}
\providecommand{\BIBentryALTinterwordspacing}{\spaceskip=\fontdimen2\font plus
\BIBentryALTinterwordstretchfactor\fontdimen3\font minus \fontdimen4\font\relax}
\providecommand{\BIBforeignlanguage}[2]{{%
\expandafter\ifx\csname l@#1\endcsname\relax
\typeout{** WARNING: IEEEtran.bst: No hyphenation pattern has been}%
\typeout{** loaded for the language `#1'. Using the pattern for}%
\typeout{** the default language instead.}%
\else
\language=\csname l@#1\endcsname
\fi
#2}}
\providecommand{\BIBdecl}{\relax}
\BIBdecl

\bibitem{wang2018action}
P.~Wang, W.~Li, C.~Li, and Y.~Hou, ``Action recognition based on joint trajectory maps with convolutional neural networks,'' \emph{Knowledge-Based Systems}, vol. 158, pp. 43--53, 2018.

\bibitem{ye2020dynamic}
F.~Ye, S.~Pu, Q.~Zhong, C.~Li, D.~Xie, and H.~Tang, ``Dynamic gcn: Context-enriched topology learning for skeleton-based action recognition,'' in \emph{Proc. ACM international conference on multimedia}, 2020, pp. 55--63.

\bibitem{journals/tip/ZhuSLZL23}
Y.~Zhu, H.~Shuai, G.~Liu, and Q.~Liu, ``Multilevel spatial‐temporal excited graph network for skeleton‐based action recognition,'' \emph{IEEE Transactions on Image Processing}, vol.~32, pp. 496--508, 2023.

\bibitem{journals/tip/ShiZCL20}
L.~Shi, Y.~Zhang, J.~Cheng, and H.~Lu, ``Skeleton‐based action recognition with multi‐stream adaptive graph convolutional networks,'' \emph{IEEE Transactions on Image Processing}, vol.~29, pp. 9532--9545, 2020.

\bibitem{journals/tip/XuYZWLJ19}
B.~Xu, H.~Ye, Y.~Zheng, H.~Wang, T.~Luwang, and Y.~Jiang, ``Dense dilated network for video action recognition,'' \emph{IEEE Transactions on Image Processing}, vol.~28, pp. 4941--4953, 2019.

\bibitem{chang2020clustering}
Y.~Chang, Z.~Tu, W.~Xie, and J.~Yuan, ``Clustering driven deep autoencoder for video anomaly detection,'' in \emph{Proc. European conference on computer vision}, 2020, pp. 329--345.

\bibitem{liu2016spatio}
J.~Liu, A.~Shahroudy, D.~Xu, and G.~Wang, ``Spatio-temporal lstm with trust gates for 3d human action recognition,'' in \emph{Proc. European conference on computer vision}, 2016, pp. 816--833.

\bibitem{rodomagoulakis2016multimodal}
I.~Rodomagoulakis, N.~Kardaris, V.~Pitsikalis, E.~Mavroudi, A.~Katsamanis, A.~Tsiami, and P.~Maragos, ``Multimodal human action recognition in assistive human-robot interaction,'' in \emph{Proc. IEEE international conference on acoustics, speech and signal processing}, 2016, pp. 2702--2706.

\bibitem{yue2022action}
R.~Yue, Z.~Tian, and S.~Du, ``Action recognition based on rgb and skeleton data sets: A survey,'' \emph{Neurocomputing}, vol. 512, pp. 287--306, 2022.

\bibitem{journals/tip/LinDHZ23}
W.~Lin, X.~Ding, Y.~Huang, and H.~Zeng, ``Self‐supervised video‐based action recognition with disturbances,'' \emph{IEEE Transactions on Image Processing}, vol.~32, pp. 2493--2507, 2023.

\bibitem{journals/tip/0001LZDLY19}
Z.~Tu, Y.~Yang, C.~Diao, B.~Li, M.~Yang, R.~Deng, G.~Wu, and S.~Li, ``Action‐stage emphasized spatiotemporal vlad for video action recognition,'' \emph{IEEE Transactions on Image Processing}, vol.~28, pp. 2799--2812, 2019.

\bibitem{chen2014improving}
C.~Chen, R.~Jafari, and N.~Kehtarnavaz, ``Improving human action recognition using fusion of depth camera and inertial sensors,'' \emph{IEEE Transactions on Human-Machine Systems}, vol.~45, pp. 51--61, 2014.

\bibitem{xu2019semisupervised}
Z.~Xu, R.~Hu, J.~Chen, C.~Chen, J.~Jiang, J.~Li, and H.~Li, ``Semisupervised discriminant multimanifold analysis for action recognition,'' \emph{IEEE transactions on neural networks and learning systems}, vol.~30, pp. 2951--2962, 2019.

\bibitem{journals/tip/GuanYHFL24}
S.~Guan, X.~Yu, W.~Huang, G.~Fang, and H.~Lu, ``Dmmg: Dual min‐max games for self‐supervised skeleton‐based action recognition,'' \emph{IEEE Transactions on Image Processing}, vol.~33, pp. 395--407, 2024.

\bibitem{journals/tip/MyungSXW24}
W.~Myung, N.~Su, J.~Xue, and G.~Wang, ``Degcn: Deformable graph convolutional networks for skeleton‐based action recognition,'' \emph{IEEE Transactions on Image Processing}, vol.~33, pp. 2477--2490, 2024.

\bibitem{shahroudy2016ntu}
A.~Shahroudy, J.~Liu, T.-T. Ng, and G.~Wang, ``Ntu rgb+ d: A large scale dataset for 3d human activity analysis,'' in \emph{Proc. IEEE conference on computer vision and pattern recognition}, 2016, pp. 1010--1019.

\bibitem{zhang2019comprehensive}
H.-B. Zhang, Y.-X. Zhang, B.~Zhong, Q.~Lei, L.~Yang, J.-X. Du, and D.-S. Chen, ``A comprehensive survey of vision-based human action recognition methods,'' \emph{Sensors}, vol.~19, p. 1005, 2019.

\bibitem{xu2022skeleton}
L.~Xu, C.~Lan, W.~Zeng, and C.~Lu, ``Skeleton-based mutually assisted interacted object localization and human action recognition,'' \emph{IEEE Transactions on Multimedia}, vol.~25, pp. 4415--4425, 2022.

\bibitem{li2023sar}
J.~Li, T.~Le, and E.~Shlizerman, ``Al-sar: Active learning for skeleton-based action recognition,'' \emph{IEEE Transactions on Neural Networks and Learning Systems}, 2023.

\bibitem{mnih2015human}
V.~Mnih, K.~Kavukcuoglu, D.~Silver, A.~A. Rusu, J.~Veness, M.~G. Bellemare, A.~Graves, M.~Riedmiller, A.~K. Fidjeland, G.~Ostrovski \emph{et~al.}, ``Human-level control through deep reinforcement learning,'' \emph{Nature}, vol. 518, no. 7540, pp. 529--533, 2015.

\bibitem{ke2017new}
Q.~Ke, M.~Bennamoun, S.~An, F.~Sohel, and F.~Boussaid, ``A new representation of skeleton sequences for 3d action recognition,'' in \emph{Proc. IEEE conference on computer vision and pattern recognition}, 2017, pp. 3288--3297.

\bibitem{du2015hierarchical}
Y.~Du, W.~Wang, and L.~Wang, ``Hierarchical recurrent neural network for skeleton based action recognition,'' in \emph{Proc. IEEE conference on computer vision and pattern recognition}, 2015, pp. 1110--1118.

\bibitem{zhang2022distilling}
Z.~Zhang, C.~Zhou, and Z.~Tu, ``Distilling inter-class distance for semantic segmentation,'' \emph{arXiv preprint arXiv:2205.03650}, 2022.

\bibitem{journals/tip/ZhangSHS18}
J.~Zhang, H.~P. Shum, J.~Han, and L.~Shao, ``Action recognition from arbitrary views using transferable dictionary learning,'' \emph{IEEE Transactions on Image Processing}, vol.~27, pp. 4709--4723, 2018.

\bibitem{journals/tip/ZhangCLWDo16}
Y.~Zhang, L.~Cheng, J.~Wu, J.~Cai, M.~N. Do, and J.~Lu, ``Action recognition in still images with minimum annotation efforts,'' \emph{IEEE Transactions on Image Processing}, vol.~25, pp. 5479--5490, 2016.

\bibitem{zhang2025}
Z.~Zhang, L.~G. Foo, H.~Rahmani, J.~Liu, and D.~W. Soh, ``Performing defocus deblurring by modeling its formation process,'' in \emph{Proceedings of the IEEE/CVF international conference on computer vision}, 2025.

\bibitem{tu2019action}
Z.~Tu, H.~Li, D.~Zhang, J.~Dauwels, B.~Li, and J.~Yuan, ``Action-stage emphasized spatiotemporal vlad for video action recognition,'' \emph{IEEE Transactions on Image Processing}, vol.~28, no.~6, pp. 2799--2812, 2019.

\bibitem{tu2023dtcm}
Z.~Tu, Y.~Liu, Y.~Zhang, Q.~Mu, and J.~Yuan, ``Dtcm: Joint optimization of dark enhancement and action recognition in videos,'' \emph{IEEE Transactions on Image Processing}, vol.~32, pp. 3507--3520, 2023.

\bibitem{liu2022motion}
Y.~Liu, J.~Yuan, and Z.~Tu, ``Motion-driven visual tempo learning for video-based action recognition,'' \emph{IEEE Transactions on Image Processing}, vol.~31, pp. 4104--4116, 2022.

\bibitem{si2020adversarial}
C.~Si, X.~Nie, W.~Wang, L.~Wang, T.~Tan, and J.~Feng, ``Adversarial self-supervised learning for semi-supervised 3d action recognition,'' in \emph{Proc. European conference on computer vision}, 2020, pp. 35--51.

\bibitem{lin2020ms2l}
L.~Lin, S.~Song, W.~Yang, and J.~Liu, ``Ms2l: Multi-task self-supervised learning for skeleton based action recognition,'' in \emph{Proc. ACM international conference on multimedia}, 2020, pp. 2490--2498.

\bibitem{thoker2021skeleton}
F.~M. Thoker, H.~Doughty, and C.~G. Snoek, ``Skeleton-contrastive 3d action representation learning,'' in \emph{Proc. ACM international conference on multimedia}, 2021, pp. 1655--1663.

\bibitem{Gong2022MetaAT}
J.~Gong, Z.~Fan, Q.~Ke, H.~Rahmani, and J.~Liu, ``Meta agent teaming active learning for pose estimation,'' in \emph{Proc. IEEE/CVF Conference on Computer Vision and Pattern Recognition}, 2022, pp. 11\,069--11\,079.

\bibitem{van2016deep}
H.~Van~Hasselt, A.~Guez, and D.~Silver, ``Deep reinforcement learning with double q-learning,'' in \emph{Proc. AAAI conference on artificial intelligence}, vol.~30, no.~1, 2016.

\bibitem{kulkarni2016hierarchical}
T.~D. Kulkarni, K.~Narasimhan, A.~Saeedi, and J.~Tenenbaum, ``Hierarchical deep reinforcement learning: Integrating temporal abstraction and intrinsic motivation,'' in \emph{Proc. Advances in neural information processing systems}, vol.~29, 2016.

\bibitem{qu2024llms}
H.~Qu, Y.~Cai, and J.~Liu, ``Llms are good action recognizers,'' in \emph{Proc. IEEE/CVF Conference on Computer Vision and Pattern Recognition}, 2024, pp. 18\,395--18\,406.

\bibitem{finn2017model}
C.~Finn, P.~Abbeel, and S.~Levine, ``Model-agnostic meta-learning for fast adaptation of deep networks,'' in \emph{Proc. International conference on machine learning}, 2017, pp. 1126--1135.

\bibitem{Rahmani2014HOPCHO}
H.~Rahmani, A.~Mahmood, D.~Q. Huynh, and A.~S. Mian, ``Hopc: Histogram of oriented principal components of 3d pointclouds for action recognition,'' in \emph{Proc. European Conference on Computer Vision}, 2014.

\bibitem{Wang2014CrossViewAM}
J.~Wang, X.~Nie, Y.~Xia, Y.~Wu, and S.-C. Zhu, ``Cross-view action modeling, learning, and recognition,'' in \emph{Proc. IEEE Conference on Computer Vision and Pattern Recognition}, 2014, pp. 2649--2656.

\bibitem{Shahroudy2016NTURA}
A.~Shahroudy, J.~Liu, T.-T. Ng, and G.~Wang, ``Ntu rgb+d: A large scale dataset for 3d human activity analysis,'' in \emph{Proc. IEEE Conference on Computer Vision and Pattern Recognition}, 2016.

\bibitem{journals/tip/ZhangWWQW18}
B.~Zhang, L.~Wang, Z.~Wang, Y.~Qiao, and H.~Wang, ``Real‐time action recognition with deeply transferred motion vector cnns,'' \emph{IEEE Transactions on Image Processing}, vol.~27, pp. 2326--2339, 2018.

\bibitem{zhang2024Diff}
Z.~Zhang, L.~Xu, D.~Peng, H.~Rahmani, and J.~Liu, ``Diff-tracker: Text-to-image diffusion models are unsupervised trackers,'' in \emph{European Conference on Computer Vision}.\hskip 1em plus 0.5em minus 0.4em\relax Springer, 2024.

\bibitem{yan2018spatial}
S.~Yan, Y.~Xiong, and D.~Lin, ``Spatial temporal graph convolutional networks for skeleton-based action recognition,'' in \emph{Proc. AAAI conference on artificial intelligence}, vol.~32, no.~1, 2018.

\bibitem{li2019actional}
M.~Li, S.~Chen, X.~Chen, Y.~Zhang, Y.~Wang, and Q.~Tian, ``Actional-structural graph convolutional networks for skeleton-based action recognition,'' in \emph{Proc. IEEE/CVF conference on computer vision and pattern recognition}, 2019, pp. 3595--3603.

\bibitem{wang20233mformer}
L.~Wang and P.~Koniusz, ``3mformer: Multi-order multi-mode transformer for skeletal action recognition,'' in \emph{Proc. IEEE/CVF Conference on Computer Vision and Pattern Recognition}, 2023, pp. 5620--5631.

\bibitem{zhang2025visual}
Z.~Zhang, Y.~Zhou, D.~Peng, J.-H. Lim, Z.~Tu, D.~W. Soh, and L.~G. Foo, ``Visual prompting for one-shot controllable video editing without inversion,'' in \emph{Proceedings of the Computer Vision and Pattern Recognition Conference}, 2025, pp. 7784--7794.

\bibitem{chen2021channel}
Y.~Chen, Z.~Zhang, C.~Yuan, B.~Li, Y.~Deng, and W.~Hu, ``Channel-wise topology refinement graph convolution for skeleton-based action recognition,'' in \emph{Proc. IEEE international conference on computer vision}, 2021, pp. 13\,359--13\,368.

\bibitem{foo2023unified}
L.~G. Foo, T.~Li, H.~Rahmani, Q.~Ke, and J.~Liu, ``Unified pose sequence modeling,'' in \emph{Proc. IEEE/CVF Conference on Computer Vision and Pattern Recognition}, 2023, pp. 13\,019--13\,030.

\bibitem{Zhang2017SemiSupervisedIA}
J.~Zhang, Y.~Han, J.~Tang, Q.~Hu, and J.~Jiang, ``Semi-supervised image-to-video adaptation for video action recognition,'' \emph{IEEE Transactions on Cybernetics}, vol.~47, pp. 960--973, 2017.

\bibitem{Singh2021SemiSupervisedAR}
A.~Singh, O.~Chakraborty, A.~Varshney, R.~Panda, R.~S. Feris, K.~Saenko, and A.~Das, ``Semi-supervised action recognition with temporal contrastive learning,'' in \emph{Proc. IEEE/CVF Conference on Computer Vision and Pattern Recognition}, 2021, pp. 10\,384--10\,394.

\bibitem{Chang2020ClusteringDD}
Y.~Chang, Z.~Tu, W.~Xie, and J.~Yuan, ``Clustering driven deep autoencoder for video anomaly detection,'' in \emph{Proc. European Conference on Computer Vision}, 2020.

\bibitem{Zheng2018UnsupervisedRL}
N.~Zheng, J.~Wen, R.~Liu, L.~Long, J.~Dai, and Z.~Gong, ``Unsupervised representation learning with long-term dynamics for skeleton based action recognition,'' in \emph{Proc. AAAI Conference on Artificial Intelligence}, 2018.

\bibitem{Si2020AdversarialSL}
C.~Si, X.~Nie, W.~Wang, L.~Wang, T.~Tan, and J.~Feng, ``Adversarial self-supervised learning for semi-supervised 3d action recognition,'' \emph{ArXiv}, vol. abs/2007.05934, 2020.

\bibitem{Lin2020MS2LMS}
L.~Lin, S.~Song, W.~Yang, and J.~Liu, ``Ms2l: Multi-task self-supervised learning for skeleton based action recognition,'' 2020.

\bibitem{lauri2022partially}
M.~Lauri, D.~Hsu, and J.~Pajarinen, ``Partially observable markov decision processes in robotics: A survey,'' \emph{IEEE Transactions on Robotics}, vol.~39, no.~1, pp. 21--40, 2022.

\bibitem{Hasselt2015DeepRL}
H.~V. Hasselt, A.~Guez, and D.~Silver, ``Deep reinforcement learning with double q-learning,'' in \emph{Proc. AAAI Conference on Artificial Intelligence}, 2015.

\bibitem{ganea2018hyperbolic}
O.~Ganea, G.~B{\'e}cigneul, and T.~Hofmann, ``Hyperbolic neural networks,'' in \emph{Proc. Advances in neural information processing systems}, vol.~31, 2018.

\bibitem{qu2024lmc}
H.~Qu, X.~Hui, Y.~Cai, and J.~Liu, ``Lmc: Large model collaboration with cross-assessment for training-free open-set object recognition,'' \emph{Advances in Neural Information Processing Systems}, vol.~36, 2024.

\bibitem{roth2006margin}
D.~Roth and K.~Small, ``Margin-based active learning for structured output spaces,'' in \emph{Proc. European Conference on Machine Learning}.\hskip 1em plus 0.5em minus 0.4em\relax Springer, 2006, pp. 413--424.

\bibitem{Sener2017ActiveLF}
O.~Sener and S.~Savarese, ``Active learning for convolutional neural networks: A core-set approach,'' \emph{arXiv}, 2017.

\bibitem{Sinha2019VariationalAA}
S.~Sinha, S.~Ebrahimi, and T.~Darrell, ``Variational adversarial active learning,'' in \emph{Proc. IEEE/CVF International Conference on Computer Vision}, 2019, pp. 5971--5980.

\bibitem{Gao2019ConsistencyBasedSA}
M.~Gao, Z.~Zhang, G.-D. Yu, S.~{\"O}. Arik, L.~S. Davis, and T.~Pfister, ``Consistency-based semi-supervised active learning: Towards minimizing labeling cost,'' in \emph{Proc. European Conference on Computer Vision}, 2019.

\bibitem{Kingma2014AdamAM}
D.~P. Kingma and J.~Ba, ``Adam: A method for stochastic optimization,'' \emph{CoRR}, vol. abs/1412.6980, 2014.

\end{thebibliography}

\begin{IEEEbiography}[{\includegraphics[width=1\textwidth]{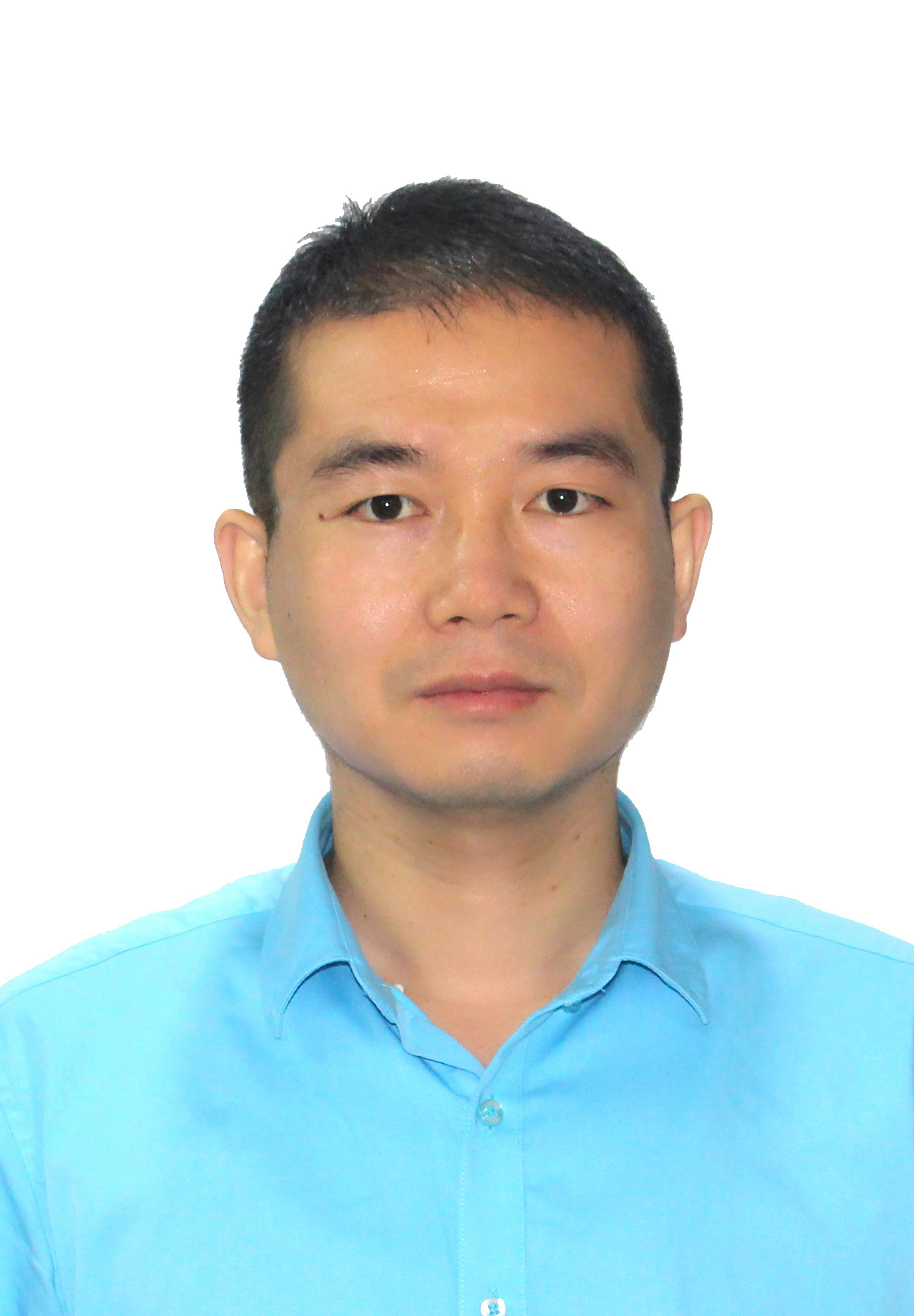}}]
{\textbf{Zhigang Tu}} (Senior Member, IEEE) received the Ph.D. degree from Wuhan University, China, in 2013, and the Ph.D. degree from Utrecht University, Netherlands, in 2015. From 2015 to 2016, he was a Postdoctoral Researcher with Arizona State University, USA. From 2016 to 2018, he was a Research Fellow with Nanyang Technological University, Singapore. 

He is currently a Professor with Wuhan University, and has co-/authored more than 80 papers in international SCI-indexed journals and conferences. His current research interests include computer vision, image processing, video analytics, machine learning, motion estimation, human action and gesture recognition, and anomaly event detection. He is the first organizer of the ACCV2020 Workshop on MMHAU, Japan. He is the Area Chair of AAAI2023/2024 and VCIP2022, an Associate Editor of the SCI-indexed journal \textit{The Visual Computer} (IF=3.5) and a Guest Editor of \textit{Journal of Visual Communications and Image Representation} (IF=2.6). He received the Best Student Paper Award at the $4^{th}$ Asian Conference on Artificial Intelligence Technology, and one of the three best reviewers awards for \textit{IEEE Transactions on Circuits and Systems for Video Technology (IEEE T-CSVT)} in 2022.
\end{IEEEbiography}

\begin{IEEEbiography}[{\includegraphics[width=1\textwidth]{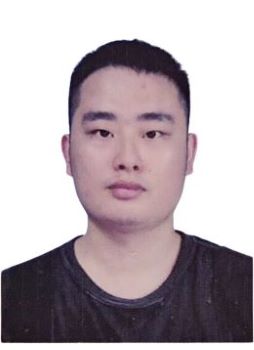}}]
{\textbf{Zhengbo Zhang}} (Graduate Student Member, IEEE) received his Bachelor's and Master's degrees in Engineering from Wuhan University. He is currently pursuing a Ph.D. in Information Systems Technology and Design at the Singapore University of Technology and Design. His current research interests are in computer vision and machine learning.
\end{IEEEbiography}

\begin{IEEEbiography}[{\includegraphics[width=1\textwidth]{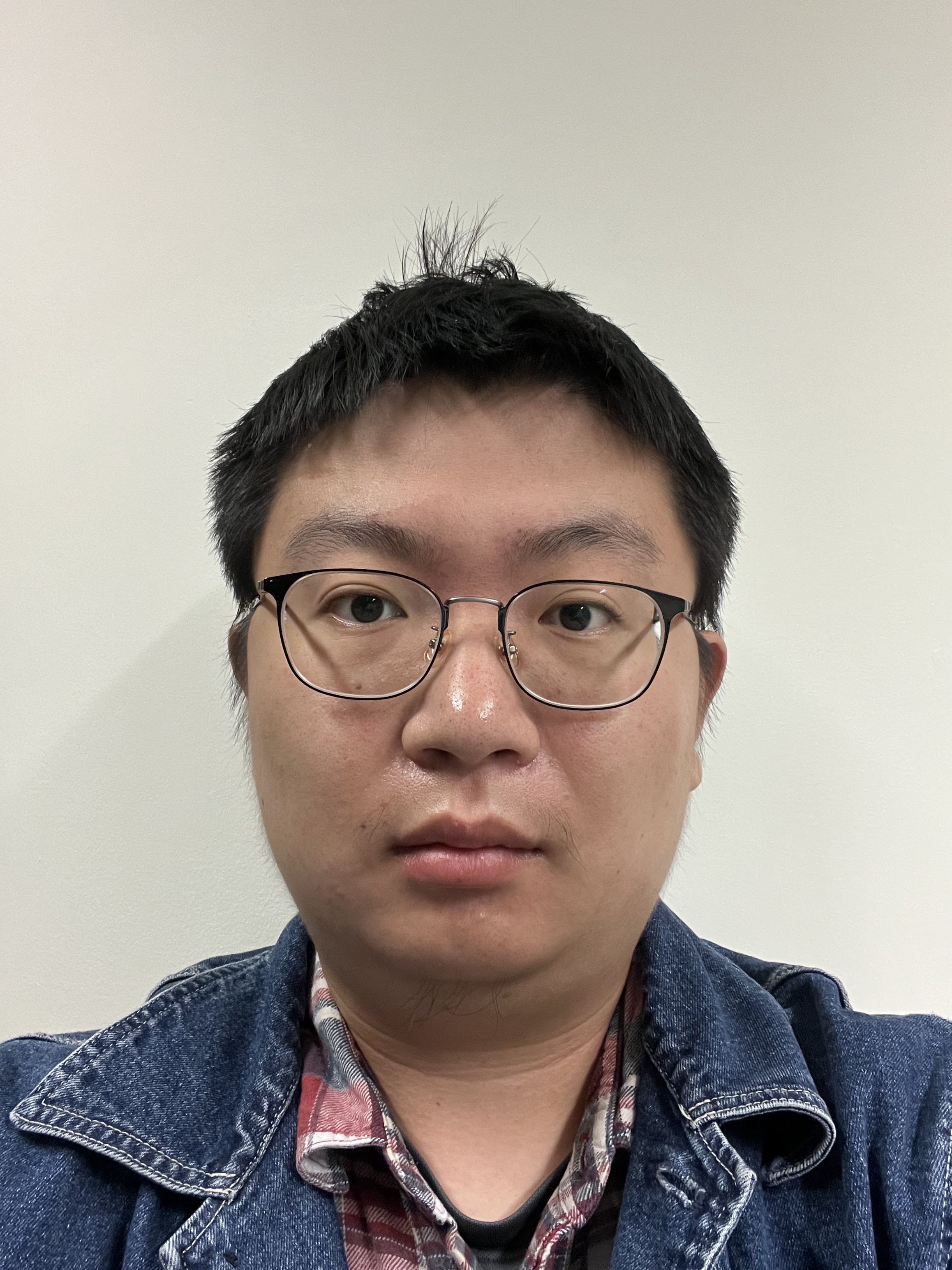}}]
{\textbf{Jia Gong}} is a researcher at the Shanghai Academy of Artificial Intelligence for Science. He received his B.Eng. degree in Optoelectronic Engineering from Chongqing University, China, and his Ph.D. in Information Systems Technology and Design from the Singapore University of Technology and Design. His research interests span human pose estimation, human mesh reconstruction, digital avatars, and active learning.
\end{IEEEbiography}

\begin{IEEEbiography}[{\includegraphics[width=1\textwidth]{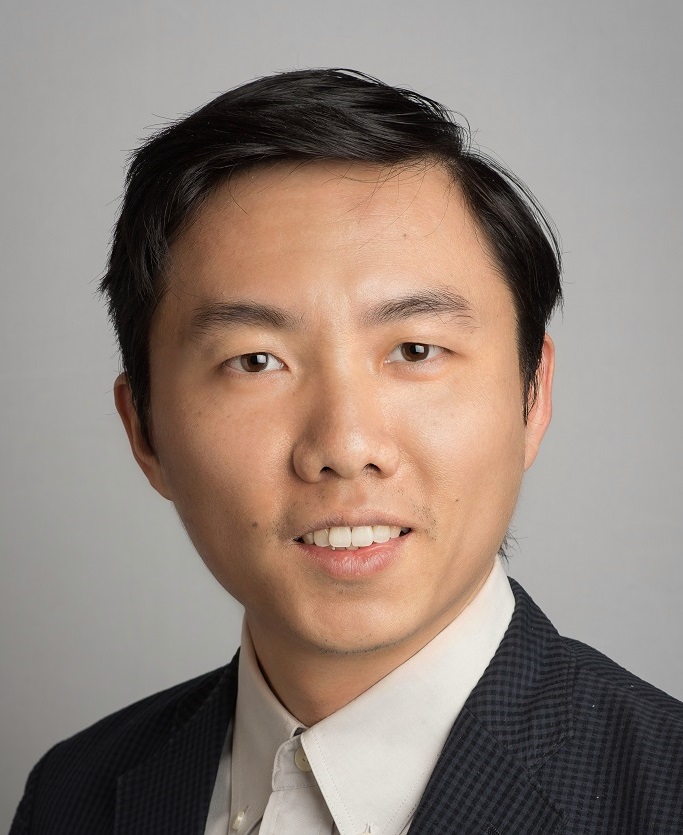}}]
{\textbf{Junsong Yuan}} (Fellow, IEEE) is Professor and Director of Visual Computing Lab at Department of Computer Science and Engineering, State University of New York at Buffalo (UB), USA. Before that he was Associate Professor (2015-2018) and Nanyang Assistant Professor (2009-2015) at Nanyang Technological University (NTU), Singapore. He obtained his Ph.D. from Northwestern University in 2009, M. Eng. from National University of Singapore in 2005, and B. Eng. from Huazhong University of Science Technology in 2002. His research interests include computer vision, pattern recognition, video analytics, large-scale visual search and mining. He received Best Paper Award from IEEE Trans. on Multimedia, Nanyang Assistant Professorship from NTU, and Outstanding EECS Ph.D. Thesis award from Northwestern University. 

He served as Associate Editor of IEEE Trans. on Pattern Analysis and Machine Intelligence (TPAMI), IEEE Trans. on Image Process. (TIP), IEEE Trans. on Circuits and Systems for Video Tech. (TCSVT), and Senior Area Editor of Journal of Visual Communications and Image Representation. He was Program Co-Chair of IEEE Conf. on Multimedia Expo (ICME'18/2022/2024), and Area Chair for CVPR, ICCV, ECCV, and ACM MM. He was elected senator at both NTU and UB. He is a Fellow of IEEE and IAPR.
\end{IEEEbiography}

\begin{IEEEbiography}[{\includegraphics[width=1\textwidth]{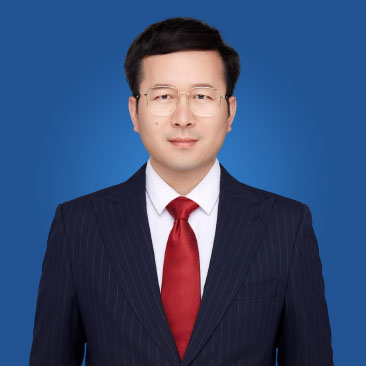}}]
{\textbf{Bo Du}} (Senior Member, IEEE) received the Ph.D. degree in photogrammetry and remote sensing from the State Key Laboratory of Information Engineering in Surveying, Mapping and Remote Sensing, Wuhan University, Wuhan, China, in 2010. He is a Professor with the School of Computer Science, Wuhan University. He has over 80 research articles published in the journals of IEEE Transactions on Pattern Analysis and Machine Intelligence (TPAMI), IEEE Transactions on Image Processing (TIP), IEEE Transactions on Geoscience and Remote Sensing (TGRS), ISPRS Journal of Photogrammetry and Remote Sensing, etc. More than 30 of them are ESI hot articles or highly cited articles. His major research interests include pattern recognition, hyperspectral image processing, machine learning, and signal processing.
\end{IEEEbiography}

\vfill

\end{document}